%% file: 0-TripleAD.tex
\documentclass[lettersize,journal]{IEEEtran}
\usepackage{amsmath,amsfonts}
\usepackage{algorithmic}
\usepackage{algorithm}
\usepackage{array}
\usepackage[caption=false,font=footnotesize,labelfont=rm,textfont=rm]{subfig}
\usepackage{textcomp}
\usepackage{stfloats}
\usepackage{url}
\usepackage{verbatim}
\usepackage{graphicx}
\usepackage{cite}
\usepackage{booktabs}
\usepackage{tabularx}
\usepackage{multirow}
\usepackage{epstopdf}
\usepackage{enumitem}
\usepackage{footnote}
\usepackage{hyperref}
\usepackage[linesnumbered, ruled, vlined, algo2e]{algorithm2e}
\usepackage{threeparttable} 
\usepackage{color}
\usepackage{etoolbox}
\usepackage{hyperref}
\hypersetup{hypertex=true, colorlinks=true, linkcolor=blue, anchorcolor=blue, citecolor=blue, urlcolor=blue}

\makeatletter
\newcommand*{\citemaincolor}{blue} 
\newcommand*{\citebracketcolor}{blue} 

\renewcommand{\@cite}[2]{%
  \textcolor{\citebracketcolor}{[}%
  \textcolor{\citemaincolor}{#1}%
  \if@tempswa , #2\fi
  \textcolor{\citebracketcolor}{]}%
}
\makeatother

\newcommand{\modelname}{TripleAD}
\newcommand{\M}{TripleAD}

\begin{document}

\title{Reconciling Attribute and Structural Anomalies for Improved Graph Anomaly Detection}

\author{{Chunjing Xiao, Jiahui Lu, Xovee Xu, Fan Zhou, Tianshu Xie, Wei Lu, and Lifeng Xu}

\thanks{Manuscript received 28 May 2024; revised 30 January 2025. \textit{(Corresponding author: Wei Lu and Lifeng Xu.)} }
\thanks{ Chunjing Xiao and Jiahui Lu are with the School of Computer and Information Engineering, Henan University, Kaifeng 475004, China (email: chunjingxiao@gmail.com, lujh@henu.edu.cn).}%
\thanks{ Xovee Xu and Fan Zhou are with the School of Information and Software Engineering, University of Electronic Science and Technology of China, Chengdu 610054, China (email: xovee@std.uestc.edu.cn, fan.zhou@uestc.edu.cn). }%
\thanks{ Tianshu Xie is with the Yangtze Delta Region Institute (Quzhou), University of Electronic Science and Technology of China, and also with the Quzhou Affiliated Hospital of Wenzhou Medical University, Quzhou People's Hospital, Quzhou 324000, China (email: tianshuxie@std.uestc.edu.cn).}
\thanks{ Wei Lu and Lifeng Xu are with the Quzhou Affiliated Hospital of Wenzhou Medical University, Quzhou People's Hospital, Quzhou 324000, China. (email: luwei@wmu.edu.cn, qz1109@wmu.edu.cn).).}%
}

\markboth{Journal of \LaTeX\ Class Files,~Vol.~14, No.~8, August~2021}%
{Shell \MakeLowercase{\textit{et al.}}: A Sample Article Using IEEEtran.cls for IEEE Journals}


\maketitle

\begin{abstract}
 Graph anomaly detection is critical in domains such as healthcare and economics, where identifying deviations can prevent substantial losses. However, existing unsupervised  approaches strive to learn a single model capable of detecting both attribute and structural anomalies. Whereas, they confront the Tug-Of-War problem between two distinct types of anomalies, resulting in sub-optimal performance. This work presents TripleAD, a mutual distillation-based Triple-channel graph Anomaly Detection framework. It includes three estimation modules to identify the attribute, structural, and mixed anomalies while mitigating the interference between different types of anomalies. In the first channel, we design a multi-scale attribute estimation module to capture extensive node interactions and ameliorate the over-smoothing issue. To better identify structural anomalies, we introduce a link-enhanced structure estimation module in the second channel that facilitates information flow to topologically isolated nodes. The third channel is powered by an attribute-mixed curvature, a new indicator that encapsulates both attribute and structural information for discriminating mixed anomalies. Moreover, a mutual distillation strategy is introduced to encourage communication and collaboration between the three channels. Extensive experiments demonstrate the effectiveness of the proposed TripleAD model against strong baselines. 
\end{abstract}


\input{1-intro}

\input{2-problem}

\input{3-method}

\input{4-experiment}

\input{5-discuss}

\input{6-relate}

\input{7-conclusion}

 

\bibliographystyle{IEEEtran}
\bibliography{IEEEbib}
\vspace{-1.2cm}

\end{document}

%% file: 1-intro.tex
\section{Introduction}

As the Internet rapidly evolves, a growing array of anomalies is causing significant disruptions and losses~\cite{liu2024generalized,zhang2024self}. In social welfare, fraudulent medical insurance claims burden healthcare systems~\cite{ma2023fighting,wu2020comprehensive}. In economic systems, financial fraud harms businesses, investors, and consumers~\cite{luo2022comga,branco2020interleaved}. On social media, fake news distorts public perception, triggering panic and confusion~\cite{nguyen2020fang,yu2018netwalk}. By modeling entities as nodes and their interactions as edges, graph anomaly detection identifies fraudulent or abnormal nodes deviating from the norm~\cite{liu2022benchmarking,xiao2024counterfactual}, offering valuable applications to address these security and economic threats.

In real-world scenarios, graph anomalies primarily manifest in three forms: attribute anomalies, structural anomalies, and mixed attribute-structural anomalies~\cite{li2017radar, zhu2020mixedad}. The attribute anomalies usually have a normal neighborhood topology but abnormal node attributes~\cite{liu2021anomaly}, whereas structural anomalies display normal node attributes but possess an atypical neighborhood topology. Mixed anomalies are characterized by irregularities in both attributes and structure. Since collecting ground-truth anomaly node labels is prohibitively expensive, the main challenge of graph anomaly detection lies in identifying anomalies in an unsupervised manner~\cite{ma2021comprehensive}. 

To address this challenge, existing efforts have been devoted to design unsupervised graph neural networks (GNNs) for graph anomaly detection, which primarily fall into two categories: reconstruction-based methods \cite{ding2019deep,huang2021hybrid,peng2020deep} and contrast-based methods~\cite{zheng2021generative, zhang2022reconstruction, duan2023graph}. Reconstruction-based methods focus on learning node representations by reconstructing node attributes and structure, identifying anomalies through reconstruction errors. On the other hand, contrast-based methods aim to learn distinguishable node representations by discriminating the agreements between nodes and their neighborhoods.

\begin{figure}[t]
  \centering
  \includegraphics[width=0.442\textwidth]{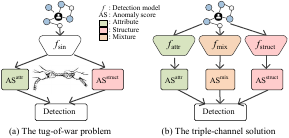}
  \vspace{-0.2cm}
  \caption{The tug-of-war problem between different anomalies. 
    (a) A single model $f_\text{sin}$ is faced with the tug-of-war issue when detecting attribute anomalies and structural anomalies simultaneously. When simultaneously detecting both types of anomalies, $f_\text{sin}$ has to compromise the performance of both tasks. 
    (b) This problem can be addressed by using three dedicated channels for different types of anomalies.}
   \vspace{-0.4cm}
  \label{fig:TugOfWar}
\end{figure}

Nevertheless, a significant limitation of current methods is their reliance on a single, unified model to detect both attribute and structural anomalies. Specifically, identifying different types of anomalies at the same time may lead to a ``tug-of-war'' problem, where the optimization for one anomaly type may interfere or diminish the optimization of the other, ultimately degrading overall performance. 
Given that attribute anomalies and structural anomalies represent inherently different types of data, the endeavor to learn distinct representations using a single model can readily induce interference between the two tasks~\cite{hadsell2020embracing, chen2023octavius}.
Recent studies also highlighted the interference between attributes and structure as a key factor in the performance decline of GNNs~\cite{yang2022graph, wang2020gcn}. An illustration of this interference problem is depicted in \figurename~\ref{fig:TugOfWar}.

Inspired by the divide-and-conquer strategy widely applied in vision~\cite{tian2021divide, wei2020component} and natural language understanding~\cite{chen2023octavius, gidiotis2020divide}, we propose to decompose the graph anomaly detection into three sub-tasks to overcome the ``tug-of-war'' problem. Specifically, 
we propose a novel Triple-channel graph Anomaly Detection framework (\M) powered by mutual distillation.
In \M, we design three distinct channels to 
identify the attribute, structural, and mixed anomalies, respectively.
Moreover, these channels engage in mutual 
cooperation
through teacher-student distillation, enhancing training effectiveness and inter-channel learning. This strategy not only mitigates the interference problem but also improves the detection performance for different types of anomalies. 

\textit{Attribute Channel}. 
This channel employs a multi-scale attribute estimation module to reconstruct a target node's masked attributes using its neighbors' attribute and structural information. By facilitating attribute propagation across multiple scales, it effectively captures extensive node interactions for attribute reconstruction. The attribute channel relieves the over-smoothing issue inherent in the multi-layer message passing of GNNs and boosts the detection of attribute anomalies.

\textit{Structure Channel}. We corrupt a target node's structure by masking its edges and try to reconstruct the masked edges utilizing both attribute and neighborhood information. We devise a link-enhanced structure estimation module that generates an enhanced graph to promote information propagation to the isolated nodes. This channel encourages node information sharing and improves the detection of structural anomalies.

\textit{Mixed Channel}. We propose a curvature-based mixture estimation module, which introduces a new indicator -- attribute-mixed curvature -- to encapsulate both attribute and structural information for detecting mixed anomalies. The dual-focus indicator simultaneously considers the strength of pairwise node connections and the degree of attribute similarity. 
By reconstructing the attribute-mixed curvature, the mixed channel captures anomalous patterns in both node attributes and structure, facilitating the detection of mixed anomalies.

\textit{Mutual Distillation}. 
We propose a mutual distillation module to promote knowledge exchange between different channels. It incorporates a triplet distillation loss to harness the complementary strengths of the attribute, structural, and mixed channels. For the attribute and structure estimation modules, we treat each other as teacher and student, and distill knowledge from the teacher to the student. For the mixture estimation module, we consider both attribute and structure estimation modules as teachers to guide the reconstruction of the attribute-mixed curvature. This strategy facilitates a rich knowledge transfer, enhancing the detection capabilities.

Our contributions are summarized as follows:
\begin{itemize}[leftmargin=*]
    \item We propose a new triple-channel graph anomaly detection framework, \modelname, which contains three distinct estimation modules for detecting attribute, structure and mixed anomalies, mitigating the cross-anomaly interference. 
    
    \item We design a multi-scale attribute estimation module, leveraging augmented views at varying propagation scales to alleviate GNN over-smoothing and boost attribute anomaly detection. 
    
    \item We devise a link-enhanced structure estimation module, which generates an enhanced graph to enable effective message passing to isolated nodes and enhances structure anomaly detection. 
    
    \item We propose a curvature-based mixture estimation module, which introduces an attribute-mixed curvature to express the mixture of attribute and structure information for mixed anomaly detection.
    
    \item We present a mutual distillation module to prompt knowledge exchange between three anomaly detection channels via teacher-student distillation. 
    
\end{itemize}
Extensive experiments demonstrate the effectiveness of proposed \M~framework in comparison to strong baselines.

%% file: 2-problem.tex
\section{Preliminaries}
\label{sec:Preliminaries}

\subsection{Problem Statement}
\label{subsec:Problem}
Now we formalize the graph anomaly detection task. For an attributed graph $\mathcal{G} = (\mathcal{V}, \mathcal{E}, \mathbf{X})$, where $\mathcal{V} = \{v_{1}, v_{2}, ..., v_{N}\}$ is a collection of nodes, $\mathcal{E} = \{e_1, e_2, \ldots, e_M\}$ is a set of edges, and $\mathbf{X} = \{x_1, x_2, \ldots, x_N\}$ is the attribute matrix of $N$ nodes. We have the adjacency matrix $\mathbf{A} \in \mathbb{R}^{N \times N}$ containing the structural information of both $\mathcal{V}$ and $\mathcal{E}$, where $\mathbf{A}_{ij} = 1$ means there exists a edge between nodes $v_i$ and $v_j$, otherwise $\mathbf{A}_{ij} = 0$. In this context, the task of a graph anomaly detection model is to rank all the nodes based on computed anomaly scores, thereby placing nodes with significant deviations from the majority of reference nodes at higher positions.

\begin{figure}[t]
    \includegraphics[width=\linewidth]{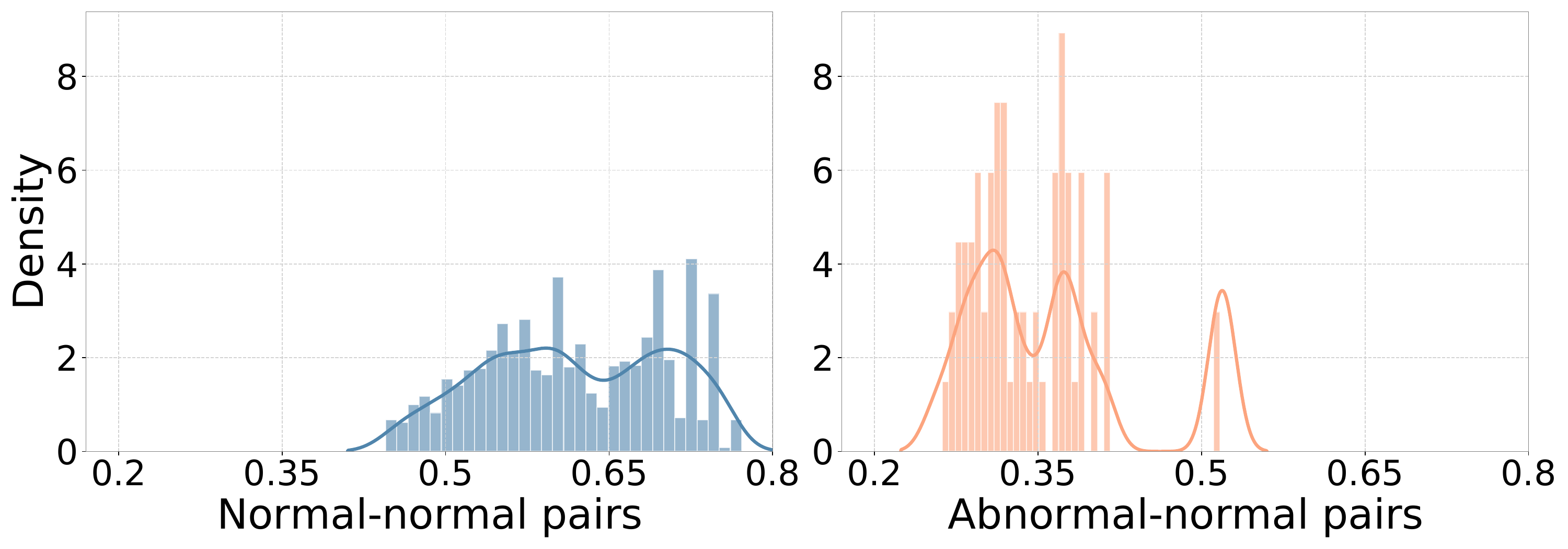}
    \vspace{-0.3cm}
    \caption{
    Curvature Distributions on CiteSeer. Normal-normal pairs (left) exhibit higher curvature values compared to normal-abnormal pairs (right), highlighting the potential of curvature as an indicator for anomaly detection.}
    \label{Fig:Curvature}
    \vspace{-0.3cm}
\end{figure}

\subsection{Graph Curvature}
\label{subsec:curvature}

Graph curvature quantifies the strength of interaction and overlap between the neighbors of a pair of nodes~\cite{ollivier2009ricci, ye2019curvature, guo2021learning}. 
Given a pair of nodes $(v_i, v_j)$, its curvature is defined as:
\begin{equation} \label{equ:curvature}
    \kappa (i, j) =1-\dfrac{W\left(\mathbf{m}_{i}, \mathbf{m}_{j}\right)}{\text{dist}(i, j)},     
\end{equation}
where $\text{dist}(,)$ is the graph distance between $v_i$ and $v_j$ and $W(, )$ refers to the Wasserstein distance between the two probability distributions, i.e., the minimum average traveling distance via any transportation plan. $\mathbf{m}_{i}$ and $\mathbf{m}_{j}$ denote the probability distribution of node $v_i$ and node $v_j$, respectively.
For node $v_i$ with degree $k$, its probability distribution is computed as:
\begin{equation}
    \mathbf{m}_{i}[x]=\left\{\begin{array}{ll}
        \alpha, & \text { if } x = i \\
        (1-\alpha) / k, & \text { if } x \in \mathcal{N}(i) \\
        0, & \text { otherwise }
        \end{array}\right.
\end{equation}
where $x = 1, ..., N$ indicates the component index of the distribution vector, $\alpha$ is a coefficient within $[0, 1]$ and $\mathcal{N}(i)$ denotes the neighbors of $v_i$.

In general, the graph curvatures of pairwise nodes with the same category tend to be higher than those with different categories. This is because nodes with the same category are inclined to share more interactions (i.e., common neighbors)~\cite{ye2019curvature, li2022curvature}. 
Correspondingly, normal-normal node pairs will have larger curvature values than normal-abnormal node pairs. 
Also, compared to other structural measures (e.g., common neighbor number and personalized PageRank), graph curvature can effectively captures abnormal connections between different communities~\cite{li2022curvature}, aligning well with the anomaly characteristics for anomaly detection.
Hence, curvature can serve as a potential indicator for anomaly detection.

To demonstrate its effectiveness, we visualize the normalized curvature distributions of normal-normal and normal-abnormal pairs on CiteSeer dataset. As depicted in Figure \ref{Fig:Curvature}, it is evident that normal-normal pairs tend to exhibit higher curvature values, primarily concentrated between 0.4 and 0.8. Conversely, normal-abnormal pairs exhibit lower curvatures, with the majority falling between 0.27 and 0.4. This observation supports the utilization of graph curvature as an effective metric for distinguishing anomalous nodes.

%% file: 3-method.tex
\section{Methodology}

We begin by providing an overview of \M~and subsequently present the details of the three channel designed for detecting attribute, structural, and mixed anomalies. At last, we illustrate the processes of inter-channel mutual distillation and model training.

\begin{figure}[t]
\centering
\includegraphics[width=0.48\textwidth]{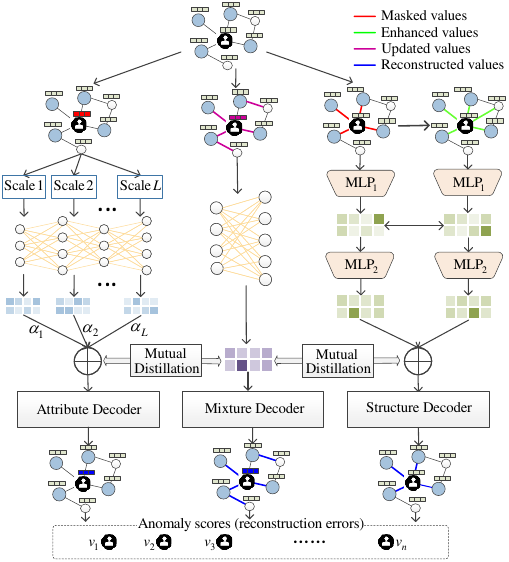}
\caption{Overview of the \modelname~framework. The three channels, the multi-scale attribute (left), the link-enhanced structure (right), and the curvature-based mixture (middle) estimation modules, are designed to identify anomalies from attribute, structural, and mixed perspectives, respectively. Furthermore, these modules interact and collaborate through our designed mutual distillation mechanism to ensure robust anomaly detection.}
\label{fig:FigTeacherOverall}
\vspace{-0.4cm}
\end{figure}

\subsection{Overview of the \modelname~Model}

We propose a mutual distillation-based Triple-channel graph Anomaly Detection framework (\M) to address the interference problem between the detection of attribute and structural anomalies. As shown in \figurename~\ref{fig:FigTeacherOverall}, \M~is composed of three channels: the multi-scale attribute estimation module (left channel), the link-enhanced structure estimation module (right channel), and the curvature-based mixture estimation module (middle channel). These three channels distinguish anomalies from attribute, structural, and mixed perspectives. Moreover, a mutual distillation strategy is proposed to promote communication and collaboration between the three channels. 
We note that the three channels are sequentially optimized (attribute$\rightarrow$structure$\rightarrow$mixture) such that they will not interfere each other and thus the ``tug-of-war'' dilemma is circumvented. This sequential optimization allows each module to focus on learning the optimal parameters for its specific task (i.e., detecting a particular type of anomaly), effectively avoiding interference from other tasks.
Before the model training, the structure module is pre-trained  for mutual distillation.

The multi-scale attribute estimation module takes the graph with masked attributes as input and produces the reconstructed attributes.
To capture both global (long-range interactions) and local information without the over-smoothing problem, this module generates multiple augmented views based on different feature propagation scales. Then, the augmented views are embedded into representations using the encoder network. At last, these representations are combined with the attention mechanism and fed into the decoder to reconstruct the node attributes.

The link-enhanced structure estimation module takes the graph with masked structure as input and produces the reconstructed structure. To enable information propagation on the target node which is isolated after masking its edges, this module first connects nodes sharing similar semantics to generate an enhanced graph. Then the original and enhanced graphs are fed into the graph diffusion-based propagation and MLP-based transformation to compute node representations. Also, the consistency loss between these two graphs is applied to enhance the model performance. At last, the representations are combined and fed into the decoder to reconstruct the structure.

The curvature-based mixture estimation module takes a graph with the attribute-mixed curvature as input and produces the reconstructed attribute-mixed curvature. To efficiently capture both attribute and structure information, this module introduces the attribute-mixed curvature to reflect both the strength of the structural connection and the attribute similarity degree. Further, this module reconstructs the attribute-mixed curvature to identify mixed anomalies.

The mutual distillation scheme connects three channels -- the attribute, structure and curvature-based mixture modules -- to further boost their performance. 
This scheme employs the triplet distillation loss based on aligning intermediate representations between three channels, prompting one to help the other for final anomaly detection.
The attribute estimation module is promoted by regarding itself as the student and the structure module as the teacher. Then, the structure module is advocated by swapping their roles. Finally, the curvature-based mixture module is advanced by regarding the other two modules as the teachers.
Hence, the anomalies are determined by combining the attribute and structure detection results.

\subsection{Multi-Scale Attribute Estimation}

An attribute-anomalous node typically exhibits a normal neighboring topology, but has abnormal attributes~\cite{liu2021anomaly}. Therefore, for a target node, we propose to mask its attributes and utilize its neighbors' attributes and topology to estimate (reconstruct) target node's masked attributes. The attribute anomalies are then identified based on the reconstruction error.

Similar to reconstruction-based methods~\cite{ding2019deep, luo2022comga}, we consider nodes with a bigger reconstruction error as anomalies -- the rationale behind this assumption is that the model is more good at memorizing the characteristics of the majority normal node. Hence, we need to accurately reconstruct the attributes for normal nodes, but not for anomalies. 
Following this idea, global information (i.e., long-range node interactions) is highly desired for effective attribute reconstruction.
The reason is that the number of normal nodes is much larger than the number of anomalies. Considering long-range node interactions means that more normal nodes are taken into account for the attribute estimation. This can let the estimated attributes to be close to the normal ones but distant from the anomalies, leading to better detection results.

However, directly stacking multiple GNN layers for learning global graph information will not only result in information distortion caused by the over-smoothing issue, but also introduce additional training burdens that hamper the model training efficacy~\cite{chen2020measuring, ding2023eliciting}.
To this end, we design a multi-scale attribute estimation module that can capture complex and rich node interactions without suffering the over-smoothing issue.
Unlike conventional GNNs, our module involves propagating features through multiple scales and using an attention mechanism to combine the extracted information. Based on the different attention scores for each-hop neighbors, our module can capture attribute information from different-hop neighbors to estimate the masked node attributes.

To capture the attribute information of the $L$-hop neighbors, we perform feature propagation with different scales to generate $L$  feature matrices $\{\bar{\mathbf{X}}^{(l)}\}^{L}_{l=1}$. Each is generated as:
\begin{equation}\label{equ:multiPropagation}
    \bar{\mathbf{X}}^{(l)} = (1-\alpha ) \bar{\mathbf{A}} \mathbf{X}^{(l-1)} + \alpha  \mathbf{X},
\end{equation}
where $\alpha$ is a restart probability and $\mathbf{X}^{(0)} = \mathbf{X}$ is the attribute matrix. $\bar{\mathbf{A}} = \tilde{\mathbf{D}}^{-\frac{1}{2}} \tilde{\mathbf{A}} \tilde{\mathbf{D}}^{-\frac{1}{2}}$ is the normalized adjacency matrix, where $\tilde{\mathbf{A}} = \mathbf{A} + \mathbf{I}$ and $\tilde{\mathbf{D}}$ is the diagonal matrix of $\tilde{\mathbf{A}}$ with the diagonal element as $\tilde{\mathbf{D}}_{i,i} = \sum_{j} \tilde{\mathbf{A}}_{i,j}$.
Here, the masked attributes of the target node are initiated as the average of its neighboring node attributes. 
Subsequently, we obtain the augmented feature matrix $\bar{\mathbf{X}}^{(l)}$ by an encoder network:
\begin{equation}\label{equ:attEncoder}
    \mathbf{Z}^{(l)}=\operatorname{ReLU}(\bar{\mathbf{X}}^{(l)} \mathbf{W}_1),
\end{equation}
where $\mathbf{Z}^{(l)} \in \mathbb{R}^{N \times h}$ is the representations of $N$ nodes in the $l$-th scale and $\mathbf{W}_1$ is a learnable parameter matrix shared for the encoding of the feature matrix. Here ${\{{\mathbf{Z}}^{(l)}\}}_{l=1}$ is learned from a local view to capture the attribute information from neighboring nodes, while ${\{{\mathbf{Z}}^{(l)}\}}^{L}_{l=2}$ is learned from a set of high-order views to capture augmented attribute information of $l$-hop distant nodes.

Through multi-scale feature propagation, we now have $L$ feature representations $\{\mathbf{Z}^{(l)}\}_{l=1}^{L}$ from $\{\bar{\mathbf{X}}^{(l)}\}^{L}_{l=1}$. Considering the difference between different hop nodes, we utilize the attention mechanism to learn representation importance:
\begin{equation}\label{equ:attention}
    (\boldsymbol{\alpha}^{(1)}, \ldots, \boldsymbol{\alpha}^{(L)}) = \text{Attention}(\mathbf{Z}^{(1)}, \ldots, \mathbf{Z}^{(L)}),
\end{equation}
where $\boldsymbol{\alpha}^{(l)} \in \mathbb{R}^{N \times 1}$ is the attention values of $\mathbf{Z}^{(l)}$. 
Further, for a target node ${v}_{i}$ with the $l$-top feature representation $\mathbf{z}_{i}^{(l)} \in \mathbb{R}^{1 \times h}$ (i.e., the $i$-th row of $\mathbf{Z}^{(l)}$), we compute its attention scores as:
\begin{equation}\label{equ:attention1}
    \omega^{(l)}_{i}=\boldsymbol{q}^{T} \cdot \tanh (\mathbf{W}^{Z} \mathbf{z}^{(l) T}_{i}+\mathbf{W}^{X} \mathbf{x}_{i}),
\end{equation}
where $\mathbf{W}^{Z} \in \mathbb{R}^{h^{\prime} \times h}$ and $\mathbf{W}^{X} \in \mathbb{R}^{h^{\prime} \times d}$ are the weight matrices, $\mathbf{x}_{i}$ represents the attribute vector of the target node, and $\boldsymbol{q} \in \mathbb{R}^{h^{\prime} \times 1}$ is the shared attention vector. The final attention weight of the target node ${v}_{i}$ is obtained by normalizing the attention values $\omega^{(l)}_{i}$ with softmax function as:
\begin{equation}  \label{equ:attention2}
  \alpha^{(l)}_{i} = \operatorname{softmax} \left(\omega^{(l)}_{i}\right) = \exp \left(\omega^{(l)}_{i}\right) / \sum_{l} \exp \left(\omega^{(l)}_{i}\right).
\end{equation}
Larger $\alpha^{(l)}_{i}$ implies that the target node ${v}_{i}$ tends to favor the attribute information of the $l$-th hop nodes. Let $\boldsymbol{\alpha}^{(l)} = \text{diag}([\alpha^{(l)}_{i}])$, we have the final representation $\mathbf{Z}$ by combining the representations at different scales:
\begin{equation}\label{equ:attentionCombine}
    \mathbf{Z} = \sum_{l}\boldsymbol{\alpha}^{(l)}\mathbf{Z}^{(l)}.
\end{equation}

Then we decode the feature representation $\mathbf{Z}$ into node attributes. Similar to the works~\cite{xie2022self, xiao2023counterfactual}, we leverage a simple fully-connected (FC) layer to reconstruct the attributes:
\begin{equation}\label{equ:attDecoder}
    \hat{\mathbf{X}} = \text{ReLU}({\mathbf{Z}}\mathbf{W}_2 + \mathbf{b}),
\end{equation}
where $\mathbf{W}_2$ is a learnable parameter matrix and $\mathbf{b}$ is the corresponding bias term.
At last, we calculate the reconstruction loss by comparing the estimated attribute matrix with the unmasked one $\mathbf{X}_\text{real}$:
\begin{equation}\label{equ:attLoss}
    \mathcal{L}_\text{attr}=\|\mathbf{X}_\text{real}-\hat{\mathbf{X}}\|_{F}^{2}. 
\end{equation}

\subsection{Link-Enhanced Structure Estimation}

Structural anomalies generally have abnormal connections with other nodes while possessing normal attributes~\cite{liu2021anomaly}.
Therefore, for a target node, we mask its links (edges) and assume its attributes are known while the edges are unknown. Then, we estimate (reconstruct) its structure and further discriminate structure anomalies based on the reconstruction errors of node links. Notably, when we reconstruct node structure, the parameters of the attribute channel are frozen such that the training of the structure estimation module is not interfered by the attribute reconstruction.

In the masked graph, the target node becomes isolated after link masking. 
During training, the isolated nodes cannot obtain knowledge from its neighboring nodes, which may weaken detection performance~\cite{liu2023learning}. To address this issue, we connect nodes sharing similar attribute semantics to generate a link-enhanced graph, which allows information propagation to the isolated nodes. Both original graph and the proposed link-enhanced graph are used to enhance the node representation learning. The learned representations can effectively capture the neighboring information and enhance structure estimation.

In particular, we employ the $k$-nearest neighbor ($k$NN) graph as the link-enhanced graph, where the isolated node is linked to semantic-similar nodes. In the link-enhanced graph, each isolated node has at least $k$ neighbors. The $k$NN graph $\mathbf{A}^{\prime}$ is constructed by using the propagated feature $\bar{\mathbf{X}}$:
\begin{equation} \label{equ:kNN}
    \mathbf{A}_{t i}^{\prime} = \left\{\begin{array}{lr}
        1 ,& v_t \in \mathcal{S}({\bar{\mathbf{X}}_i},k) \: \text{or} \: v_i \in \mathcal{S}(\bar{\mathbf{X}}_t,k),\\
        0 ,&\text { otherwise,}
        \end{array}\right.
\end{equation}
where $\mathcal{S}\left(\bar{\mathbf{X}}_{t}, k\right)$ is the set of $k$ nodes with the highest similarity to $\bar{\mathbf{X}}_{t}$. Here the propagated feature $\bar{\mathbf{X}}$ is computed based on the masked graph using graph diffusion propagation (cf. Equation~\eqref{equ:propagation2}).

Having the link-enhanced graph, we utilize graph diffusion followed by transformation to compute the node representations via two steps: the graph-based diffusion propagation and the FC-based transformation. 
Assuming $\bar{\mathbf{A}}$ and $\bar{\mathbf{A}}^{\prime}$ denote the normalized adjacency matrices of the masked and enhanced graphs, respectively, graph diffusion propagation for both graphs can be expressed as:
\begin{align} \label{equ:propagation1}
    \mathbf{X}^{(\mathbf{t}+1)} &= (1-\beta) \bar{\mathbf{A}} \mathbf{X}^{(\mathbf{t})} + \beta \mathbf{X}, \;
    \\
    \label{equ:propagation11} 
    \mathbf{X}^{{\prime}(\mathbf{t}+1)} &= (1-\beta) \bar{\mathbf{A}}^{\prime} \mathbf{X}^{(\mathbf{t})} + \beta \mathbf{X}, 
\end{align}
where $\mathbf{X}^{(0)} = \mathbf{X}$ and $\beta \in$ (0,1] is a restart probability. After $T$ iterations, we have propagated feature matrices :
\begin{equation} \label{equ:propagation2}
    \bar{\mathbf{X}} = \mathbf{X}^{(T)},\quad
    \bar{\mathbf{X}}^{\prime} = \mathbf{X}^{{\prime}(T)}.
\end{equation}

Then, we utilize two FC layers to map $\bar{\mathbf{X}}$ and $\bar{\mathbf{X}}^{\prime}$ to the intermediate representations  $\mathbf{H}_2$ and $\mathbf{H}_2^{\prime}$, respectively. The transformation of the masked and enhanced graphs are written as:
\begin{align} \label{equ:encoderMLP}
        \mathbf{H}_{1} &= \text{FC}_{1}(\bar{\mathbf{X}};\mathbf{W}_{3}) ,
        &\mathbf{H}_{2} = \text{FC}_{2}(\mathbf{H}_{1}; \mathbf{W}_{4}) ,\\
        \label{equ:encoderMLP2}
        \mathbf{H}_{1}^{\prime} &= \text{FC}_{1}(\bar{\mathbf{X}}^{\prime};\mathbf{W}_{3}) , 
        &\mathbf{H}_{2}^{\prime} = \text{FC}_{2}(\mathbf{H}_{1}^{\prime}; \mathbf{W}_{4}) ,
\end{align}
where $\mathbf{W}_{3}$ and $\mathbf{W}_{4}$ are learnable parameter matrices. 

At last, we combine the representations $\mathbf{H}_{2}$ and $\mathbf{H}_{2}^{\prime}$ from both graphs to estimate the adjacency matrix $\hat{\mathbf{A}}$:
\begin{equation}\label{equ:strDecoder}
    \begin{aligned}
        \hat{\mathbf{A}} &= f_{d}(\operatorname{AGG}\left(\mathbf{H}_{2}, \mathbf{H}_{2}^{\prime}\right)), 
    \end{aligned}
\end{equation}
where $f_{d}(\cdot)$ is the reconstruction function that estimates the adjacency matrix by performing similarity or the inner product between node representations; $\operatorname{AGG}(\cdot)$ denotes the aggregation function such as concatenation. We minimize the reconstruction error $\mathcal{L}_\text{adj}$ of the estimated adjacency matrix with the realistic one:
\begin{equation}\label{equ:strReconstructLoss}
    \mathcal{L}_\text{adj}=\|\mathbf{A}_\text{real}-\hat{\mathbf{A}}\|_{F}^{2}.
\end{equation}

However, this reconstruction objective has a notable limitation. 
Since the masked and link-enhanced graphs share model parameters, the learned node representations (especially for the isolated nodes) may be significantly different due to the difference in the input structure. 
Hence, we introduce a consistency alignment loss, which promotes the semantic consistency between the two graphs: 
\begin{equation}\label{equ:calLoss}
    \mathcal{L}_\text{cons}=\|\mathbf{H}_{1}-{\mathbf{H}_{1}^{\prime}}\|_{F}^{2}.
\end{equation}
This loss encourages the consistency between original view and link-enhanced view, helping the model extract more relevant information \textit{w.r.t.} the isolated nodes. 

During training, we combine the two losses as the overall objective of the structure estimation, with $\gamma$ a weighting coefficient:
\begin{equation}\label{equ:strLoss}
    \begin{aligned}     \mathcal{L}_\text{str}&=\mathcal{L}_\text{adj}+\gamma\mathcal{L}_\text{cons}.
    \end{aligned}
\end{equation}

\subsection{Curvature-based Mixture Estimation}

Mixed anomalies display characteristics that are anomalous \textit{w.r.t.} both attributes and structure in a mixed manner~\cite{ma2021comprehensive, zhu2020mixedad}. To distinguish mixed anomalies, we design a new indicator, termed attribute-mixed curvature, to reflect the mixed nature of attributes and structure, and then we reconstruct this indicator and discriminate mixed anomalies according to the reconstruction error.

The rationale behind this new indicator for detecting mixed anomalies is due to the following reasons:
(1) graph curvature is indicative of significant disparities between normal nodes and anomalies. It quantifies the interaction or strength of the overlap between pairs of connected nodes~\cite{ollivier2009ricci, ye2019curvature, guo2021learning}.
As confirmed in~\cite{li2022curvature}, node pairs with the same category tend to exhibit higher curvature values, attributing to a greater number of shared neighbors, whereas node pairs with different categories display lower curvature values. 
In context of anomaly detection, 
given that neighbors of both normal and abnormal nodes are predominantly normal (owing to the higher prevalence of normal nodes), abnormal nodes tend to have lower curvature values due to less overlap with their neighbors (referring to Section~\ref{subsec:curvature} for empirical analysis). Therefore, measuring curvature allows for the differentiation of abnormal nodes. 
(2) Incorporating node attributes into the curvature calculation can strengthen its effectiveness in distinguishing mixed anomalies.
We incorporate the attribute similarity degree into the curvature. Node pairs with higher attribute similarity are assigned greater curvature values -- this assignment is relatively higher for normal nodes and lower for abnormal nodes, since normal nodes tend to have higher similarity with their neighbors, which are dominated by normal nodes, while anomalous nodes have lower similarity. Consequently, this consideration decreases the curvature values of anomalies, facilitating their detection.

\noindent\textbf{Computing Attribute-Mixed Curvature.}
We compute the attribute-mixed curvature $\kappa (i, j)$ for node pair ($v_i$, $v_j$) as follow:
\begin{equation}\label{equ:attCurvature}
    \kappa (i, j) =1-\left( W \left(\hat{\mathbf{m}}_{i(j)}, \hat{\mathbf{m}}_{j(i)} \right) / d(i,j) \right), 
\end{equation}
where $d(,)$ is the graph distance between $v_i$ and $v_j$ and $W(, )$ refers to the Wasserstein distance between the two distributions.
$\hat{\mathbf{m}}_{i(j)}$ is defined as a probability distribution vector of node $v_i$ to $v_j$. When calculating $\hat{\mathbf{m}}_{i(j)}$, instead of considering all the neighbors equally, we divide the neighbors into two groups -- common and uncommon neighbors -- and use different strategies for them:
\begin{equation}\label{equ:distribution}\small
    \hat{\mathbf{m}}_{i(j)}[x] = \left \{\begin{array}{ll}
        \delta , & \text{if} \; x=i,\\ \vspace{0.1cm}
        (1-\delta + \dfrac{\mathbf{S}'_{ij}}{|\mathcal{N}_{i}\cap \mathcal{N}_{j} |} ) \dfrac{1}{k}, &\text{if} \; x \in \mathcal{N}_{i}\cap \mathcal{N}_{j},\\ 
        (1-\delta - \dfrac{\mathbf{S}'_{ij}}{|\mathcal{N}_{i} - \mathcal{N}_{j} |}  ) \dfrac{1}{k}, &\text{if} \; x \in \mathcal{N}_{i}- \mathcal{N}_{j},\\
        0, & \text{otherwise},
    \end{array}
      \right.
\end{equation}
where $x = 1, ..., N$ is the index of the distribution vector $\hat{\mathbf{m}}_{i(j)}$, $\delta \in [0, 1]$ is a coefficient, and $\mathcal{N}_{i} \cap \mathcal{N}_{j}$ represents the common neighbor set of $v_i$ and $v_j$, while $\mathcal{N}_{i}-\mathcal{N}_j$ refers to the difference between the neighbor sets of $v_i$ and $v_j$. $\mathbf{S}'_{ij} = \mathbf{S}_{ij}/(1-\delta)$ is the normalized attribute similarity between $v_i$ and $v_j$.

\noindent\textbf{Updating Adjacent Matrix.}
Further, we normalize the curvature values from 0 to 1 using a monotonically increasing sigmoid function: $\kappa'(i, j) = 1/ \left(1 + \exp(-\kappa (i,j)\right)$. 
The normalized curvature values are adopted to update the adjacent matrix for reconstruction. The original adjacent matrix $\mathbf{A}_{ij}$ is updated to $\mathbf{C}_{ij}$:
\begin{equation}\label{equ:attCurvatureMatrix}
    \mathbf{C}_{ij} = \left\{\begin{array}{ll}
        1, & \text{if  } i = j,\\
        {\kappa}^{\prime}(i, j), & \text{if  } \mathbf{A}_{ij} = 1, \\
        0, & \text{otherwise}.\\
    \end{array}
    \right.
\end{equation}

\noindent\textbf{Reconstructing Attribute-Mixed Curvature.}
Based on the updated adjacent matrix and attribute matrix, we use GCN~\cite{kipf2016semi} as the encoder to embed nodes into  representations for reconstructing the attribute-mixed curvature values. 
Specifically, multiple GCN layers are adopted to aggregate node representations. Here, the $l$-th layer representation is obtained by the forward encoding process:
\begin{equation}\label{equ:attCurvatureEncoder}
    \mathbf{H}_{c}^{(l)} = \sigma(\mathbf{L} \mathbf{H}_c^{(l-1)} \mathbf{W}^{(l-1)})
\end{equation}
where $\sigma(\cdot)$ is a non-linear activation function, $\mathbf{H}_c^{(0)} = \mathbf{X}$, and $\mathbf{W}^{(l-1)}$ denotes the $l$-th learnable parameter matrix. $\mathbf{L}$  refers to the symmetrically normalized Laplacian matrix $\mathbf{L} = \tilde{\mathbf{D}}_{c}^{-1/2} \mathbf{C}\tilde{\mathbf{D}}_c^{-1/2}$, where $\tilde{\mathbf{D}}_c$ is the degree matrix of $\mathbf{C}$. 
The attribute-mixed curvature is reconstructed by the obtained representation: 
\begin{equation}\label{equ:mixDecoder}
\hat{\mathbf{C}} = f(\mathbf{H}_c^{(L)} \cdot {\mathbf{H}_c^{(L)}}^T), 
\end{equation}
where $f(\cdot)$ denotes the FC layers for reconstructing attribute-mixed curvature. At last, we compute the reconstruction error by comparing the reconstructed curvature and the original one: 
\begin{equation}\label{equ:lmix}
\mathcal{L}_\text{mix} = \|\mathbf{C}-\hat{\mathbf{C}}\|_{F}^{2}.
\end{equation}

\subsection{Mutual Distillation and Model Optimization}

Now we have the three channels, i.e., the attribute, structure and mixture estimation modules. 
These modules are highly correlated and complementary as they try to learn the representations of the same node from different aspects. 
Correspondingly, one module can be boosted by receiving knowledge from another module.
Hence, we argue that a teacher-student distillation is beneficial for knowledge transferring between different modules. However, general distillation methods focus on minimizing the difference between teacher and student~\cite{gou2021knowledge, tian2025knowledge}.
Whereas, our multi-channel method requires learning node representations that belong to the same category but also with some differences. Inspired by~\cite{oki2020triplet, boutros2022self}, we propose to employ the triplet loss of metric learning to embed the knowledge of the teacher in the output space of the student and simultaneously clarify the difference between their outputs.
This scheme enables the triple channels to mutually enhance each other to improve the overall detection performance.

Specifically, to enhance the attribute estimation module, we consider it as the student and the structure module as the teacher. 
Then, we select an anchor sample $x_a$, a positive $x_p$, and a negative $x_n$ to form a triplet $(x_a$, $x_p$, $x_n)$. 
Here, anchor $x_a$ and positive $x_p$ are the representations of the same node from the student and teacher, respectively, and negative $x_n$ is a randomly selected sample that differs from the node of $x_a$. 
Based on this triplet, the mutual distillation scheme learns representations such that the anchor-to-positive distance is relatively closer than that of the anchor-to-negative.
As a result, this objective can encourage the representations of the same node from both modules to be close but also keep some differences, meeting the requirement of the two modules with different goals and distilling knowledge from one module to enhance another.

The triplet distillation loss $\mathcal{L}_\text{attr}^d$ is defined as:
\begin{equation}
    \begin{aligned}
        \mathcal{L}_\text{attr}^{d} =&\sum_{(a, p, n) \in \Omega} \max \left(0, \left\|t\left(\boldsymbol{x}_{p}\right)-s\left(\boldsymbol{x}_{a}\right)\right\|_{2}^{2}\right. \\
        &-\left.\left\|t\left(\boldsymbol{x}_{n}\right)-s\left(\boldsymbol{x}_{a}\right)\right\|_{2}^{2} + m \right),
        \end{aligned}
        \label{equ:distillation}
\end{equation}
where $t(\cdot)$ and $s(\cdot)$ denote outputs of the teacher and student, respectively, and $m$ is a hyper-parameter that defines how far away the dissimilarities should be. 
$\Omega$ is an index set representing each corresponding anchor, positive, and negative. Since we are training the student model, the weight of the teacher model is frozen.

For the attribute module, the final training objective is the combination of the reconstruction loss and the distillation loss:
\begin{equation}
    \mathcal{L}_\text{attr}^{\text {fin}}= \eta_1 \mathcal{L}_\text{attr} + \eta_2 \mathcal{L}_\text{attr}^{d},
    \label{equ:finalLossAtt}
\end{equation}
where $\eta_1$ and $\eta_2$ are the balancing hyper-parameters.

Likewise, to boost the link-enhanced structure estimation, we swap the roles of the two modules, i.e., the attribute module as teacher and the structure module as the student. The distillation loss for the structure estimation module, denoted as $\mathcal{L}_\text{str}^d$, is derived similar to Eq.~\eqref{equ:distillation}. Consequently, the final training objective for the structure estimation module is the combination of the reconstruction loss and the distillation loss: 
\begin{equation}
    \mathcal{L}_\text{str}^{\text {{fin} }}= \eta_1 \mathcal{L}_\text{str} + \eta_2 \mathcal{L}_\text{str}^{d}.
    \label{equ:finalLossStr}
\end{equation}

At last, to advocate the curvature module, we consider both attribute and structure modules as the teachers to guide the reconstruction of the attribute-mixed curvatures.  
Similar to Eq.~\eqref{equ:distillation}, the distillation losses,  
$\mathcal{L}_\text{mix}^\text{attr}$ and $\mathcal{L}_\text{mix}^\text{str}$, are obtained when regarding the attribute and structure modules as the teachers, respectively. 
The final distillation loss for the curvature-based mixture estimation module is:
\begin{equation}\label{equ:finalLossMix}
    \mathcal{L}_\text{mix}^{\text {fin}} = \eta_1 \mathcal{L}_\text{mix} + \eta_2 \big(\mathcal{L}_\text{mix}^\text{attr} + \mathcal{L}_\text{mix}^\text{str}\big).
\end{equation}

\subsection{Anomaly Scores}

Similar to existing works~\cite{tong2011non, peng2020deep}, we distinguish anomalies from the normal ones according to the reconstruction error, i.e., the larger the reconstruction error, the more likely the objects are anomalies. This is because that the anomalies fail to conform to the patterns of the majority and hence cannot be accurately reconstructed. 
For a given node $v_i$, 
we compute its attribute, structure and attribute-mixed curvature reconstruction errors as follows:
\begin{flalign} \label{equ:attAnomalyScore}
    \text{AS}^\text{attr}(v_i) = \|\mathbf{X}_{i}-\hat{\mathbf{X}}_{i}\|^{2}, \\
\label{equ:strAnomalyScore}
    \text{AS}^\text{str}(v_i) = \|\mathbf{A}_{i}-\hat{\mathbf{A}}_{i}\|^{2}, \\
\label{equ:mixAnomalyScore}
    \text{AS}^\text{mix}(v_i) = \|\mathbf{C}'_{i}-\hat{\mathbf{C}}'_{i}\|^{2},
\end{flalign}
where $\mathbf{C}'_{i}$ and $\hat{\mathbf{C}}'_{i}$ denote the original and reconstructed node attribute-mixed curvatures, which are computed by averaging the attribute-mixed curvatures (i.e., $\mathbf{C}_{ij}$ and $\hat{\mathbf{C}}_{ij}$) of node pairs involving this node.

After combining these three types of reconstruction errors, the final anomaly score of node $v_i$ is calculated as follows:
\begin{equation}\label{equ:anomalyScores}
    \text{AS}(v_i) = \lambda_1 \cdot \text{AS}^\text{attr}(v_i) +  \lambda_2\cdot \text{AS}^\text{str}(v_i) + \lambda_3 \cdot \text{AS}^\text{mix}(v_i),
\end{equation}
where $\lambda_1, \lambda_2 \text{ and } \lambda_3$ are three balance hyper-parameters. The anomalous nodes are distinguished by the anomaly scores, i.e., 
nodes with higher scores are more likely to be anomalies.

%% file: 4-experiment.tex
\section{Experiment}
\label{sec:experiment}
In this section, we carry out performance evaluations to demonstrate the efficacy of \modelname~for graph anomaly detection, as well as the ablation study and sensitivity analysis.

\begin{table}[t]
  \footnotesize
  \caption{Descriptive Statistics of Datasets} 
  \vspace{-0.3cm}
  \setlength{\tabcolsep}{3.5pt}
  \centering
  \begin{tabular}{llrrrr}
  \toprule
     & dataset  & \# nodes & \# edges & \# features & \# anomalies\\
  \midrule
  
   \multirow{2}{*}{Ground-truth} & Amazon    & 1,418  & 3,695  & 28  & 28  \\
    & YelpChi   & 45,954  & 3,846,979  & 32  & 6,677\\
  \midrule
    \multirow{3}{*}{Injected} & CiteSeer & 3,327  & 4,732  & 3,703  & 150  \\
     & ACM & 16,484  & 71,980  & 8,337  & 600 \\
    & Flickr & 7,575  & 239,738  & 12,047  & 450 \\
  \bottomrule
  \end{tabular}
  \label{tab:datasets}
  \vspace{-0.5cm}
\end{table}

\subsection{Experimental Settings}

\noindent\textbf{Datasets.}
We conduct experiments on two types of datasets, whose details are shown in Table \ref{tab:datasets}.
We split the datasets into training, validation, and test sets with a ratio of 6:2:2, and hyperparameters are tuned based on the validation set.
\begin{itemize}
  \item \textit{Ground-truth anomaly graphs}: 
  Amazon~\cite{luo2022comga} is a co-purchase network and YelpChi~\cite{tang2022rethinking} is a transaction network, both of which have ground-truth labels of the anomalies and contain the three types of anomalies.
  \item \textit{Injected anomaly graphs}: CiteSeer~\cite{liu2021anomaly} and {ACM}~\cite{ding2021inductive} are two citation networks, and {Flickr}~\cite{ding2019deep} is a social network, with injected anomaly labels.
   Attribute, structural and mixed anomalies are injected into these three datasets using injection ways of previous studies~\cite{ding2019deep, liu2021anomaly}.
\end{itemize}

\noindent\textbf{Baselines.}
We compare our model {\modelname} with various baselines from three major categories:
\begin{itemize}
  \item \textit{Traditional methods}: AMEN~\cite{perozzi2016scalable}, Radar~\cite{li2017radar} and ANOMALOUS~\cite{peng2018anomalous} employ shallow traditional techniques to implement anomaly detection.
  \item \textit{Reconstruction-based approaches}: DOMINANT~\cite{ding2019deep}, AnomalyDAE~\cite{fan2020anomalydae}, ALARM~\cite{peng2020deep}, ComGA~\cite{luo2022comga}, ADA-GAD~\cite{he2024ada} and GAD-NR~\cite{roy2024gad} perform anomaly detection via the autoencoder-based reconstruction technique. 
  \item \textit{Contrast-based methods}: CoLA~\cite{liu2021anomaly}, Sub-CR~\cite{zhang2022reconstruction}, GRADATE~\cite{duan2023graph} and FedCAD~\cite{kong2024federated} conduct anomaly detection through contrastive learning.
\end{itemize}

\noindent\textbf{Experimental Setup.}
We sequentially optimize attribute, structure, and mixture estimation modules, which prevents interference between the loss functions of the three modules and ensures that each module learns effectively.
For all experiments, we use grid search to find the optimal hyper-parameters, such as the learning rate and number of epochs. 
The grid search range for these parameters is shown in Table~\ref{tab:parameter}.
Note that for the YelpChi dataset, we utilize sparse matrices instead of dense matrices whenever possible to mitigate out-of-memory issues.
For other baselines, we retain to the settings described in the corresponding papers to report optimized results. 

\begin{table}[h]
  \centering
  \caption{Hyper-parameter search space.}
  \vspace{-0.3cm}
  \setlength{\tabcolsep}{6.5pt}
  \begin{tabular}{ll}
  \toprule
  \textbf{Parameter} & \textbf{Search Space} \\
  \midrule
  Scale $L$ & \{1, 2, 3, 4, 5, 6, 7, 8\} \\
  Propagation iteration $T$ & \{1, 2, 3, 4, 5, 6, 7, 8\} \\
  Restart probability $\alpha$ and $\beta$ & \{0.001, 0.05, 0.1, 0.2\} \\
  Hidden unit $h$ & \{32, 64, 128, 256\} \\
  Learning rate & \{0.1, 0.01, 0.001, 0.0001\} \\
  Balance parameters $\lambda_1$,$\lambda_2$,$\lambda_3$ & \{0.1, 0.2, 0.3, 0.4, 0.5, 0.6, 0.7, 0.8, 0.9\}\\
  \bottomrule
  \end{tabular}
  \label{tab:parameter}
  \vspace{-0.3cm}
  \end{table}

\subsection{Anomaly Detection Results}

We report the anomaly detection results of {\modelname} and the baselines in Table \ref{tab:detectResults}. We use widely used anomaly detection metrics including AUC-ROC curve (AROC), AUC-PR curve (APR), and Macro F1 score (F1). Specifically, we have the following observations. 

First, the \modelname~ model demonstrates superior detection performance across all evaluated datasets, outperforming the baseline models in every metric. This improvement is attributed to our model's unique approach of learning three distinct representations, enhanced through a mutual distillation scheme. This strategy effectively mitigates the challenges associated with the
interference between attributes and structure while specifically addressing the detection of mixed anomalies.

Second, the reconstruction-based methods (e.g., ADA-GAD and GAD-NR) show good performance compared to the traditional methods. This indicates their effectiveness in capturing anomalous patterns from high-dimensional attributes and complex structure. However, our method significantly outperforms them by employing the triple-channel strategy.
Note that AnomalyDAE, which adopts two modules to compute attribute and structural representations, displays inferior performance compared to our model. 
The main reason is that it fails to mitigate the interference, as it concatenates both representations to simultaneously identify different anomalies in the same way.

Third, the contrast-based methods (e.g., GRADATE and FedCAD) exhibit remarkable performance. These approaches, in particular, 
leverage the unsupervised contrastive learning, achieve excellent detection performances.
However, a notable limitation of these methods is their reliance on a singular model for detecting both attribute and structural anomalies, which may result in modality interference. In contrast, \modelname~outperforms these methods in non-trivial margins, validating our motivation for adopting a triple-channel strategy to alleviate the interference between attribute anomaly detection and structure anomaly detection. 
In addition, we calculate the runtime of our model, showing that its speed is competitive with existing methods and is well-suited for large-scale datasets.

\begin{table*}[t]
  \fontsize{7.2}{9.2}\selectfont
  \setlength{\tabcolsep}{3.0pt}
  \centering
  \caption{Performance comparison between \modelname~and baselines on the five datasets.}
  \vspace{-0.3cm}
  \begin{tabular}{lccccccccccccccc}
    \toprule
    \textbf{Dataset} & \multicolumn{3}{c}{\textbf{Amazon}}              & \multicolumn{3}{c}{\textbf{YelpChi}}            & \multicolumn{3}{c}{\textbf{CiteSeer}}    & \multicolumn{3}{c}{\textbf{ACM}}  & \multicolumn{3}{c}{\textbf{Flickr}}            \\
    \cmidrule(lr){2-4} \cmidrule(lr){5-7} \cmidrule(l){8-10} \cmidrule(l){11-13} \cmidrule(l){14-16}
    Metric & \footnotesize{AROC} & \footnotesize{APR} & \footnotesize{F1}  & \footnotesize{AROC}  & \footnotesize{APR} & \footnotesize{F1}  & \footnotesize{AROC}  & \footnotesize{APR} & \footnotesize{F1} & \footnotesize{AROC}  & \footnotesize{APR} & \footnotesize{F1}& \footnotesize{AROC}  & \footnotesize{APR} & \footnotesize{F1}   \\
    \midrule
    AMEN & 47.00  & 21.38  & 17.65  & 23.18  & 12.45  & 18.32  & 61.54  & 56.42  & 59.94  & 56.26  & 34.33  & 41.46  & 65.73  & 57.12  & 53.30  \\ 
    Radar & 58.01  & 26.38  & 20.07  & 34.16  & 13.21  & 20.24  & 67.09  & 53.51  & 66.00  & 72.47  & 35.67  & 46.12  & 73.99  & 63.68  & 57.15  \\ 
    ANOMALOUS & 60.20  & 28.80  & 22.34  & 48.59  & 13.92  & 25.52  & 63.07  & 58.78  & 67.02  & 70.38  & 38.56  & 53.76  & 74.34  & 69.27  & 59.27  \\ 
    \midrule
    DOMINANT & 62.50  & 33.18  & 28.79  & 49.32  & 15.58  & 35.86  & 82.51  & 76.90  & 46.27  & 76.01  & 50.95  & 58.08  & 74.42  & 69.35  & 59.32  \\ 
    AnomalyDAE & 53.61  & 25.07  & 23.76  & 36.13  & 13.15  & 30.56  & 72.71  & 70.35  & 58.63  & 75.13  & 48.23  & 50.28  & 75.08  & 70.31  & 58.16  \\
    ALARM & 62.09  & 34.30  & 37.16  & 41.68  & 16.90  & 37.28  & 84.31  & 78.80  & 63.07  & 78.34  & 51.13  & 54.07  & 76.04  & 70.89  & 60.80  \\ 
    ComGA & 63.07  & 41.57  & 44.73  & 49.26  & 20.64  & 40.61  & 91.67  & 82.25  & 75.38  & 84.96  & 53.11  & 58.41  & 79.91  & 63.16  & 62.71  \\ 
    ADA-GAD & {63.87}  & 41.98  & 45.13  & 49.34  & 21.46  & 41.36  & 92.74  & 83.22  & 76.13  & 85.69  & 53.75  & 58.68  & 80.26  & 64.44  & 63.12 \\ 
    GAD-NR & \underline{64.12}  & 42.07  & 46.05  & \underline{49.42}  & 21.81  & 42.03  & 93.42  & 84.67  & 78.85  & 86.27  & 54.81  & 59.06  & 80.04  & 59.83  & 62.92 \\ 
    \midrule
    CoLA & 47.26  & 47.22  & 43.39  & {49.38}  & 26.47  & 43.08  & 89.68  & 83.44  & 71.02  & 82.37  & 51.48  & 53.97  & 75.13  & 69.95  & 62.05  \\ 
    ANEMONE & 59.64  & 48.07  & 46.95  & 48.62  & 26.10  & 44.85  & 91.89  & 85.49  & 79.04  & 83.00  & 49.86  & 54.85  & 76.37  & 71.28  & 61.73  \\ 
    Sub-CR & 62.66  & 48.21  & 49.41  & 46.65  & 26.36  & 45.41  & 93.03  & 84.21  & 77.84  & 84.28  & 50.69  & 56.02  & 79.75  & 71.33  & 64.07  \\
    GRADATE & 62.79  & 49.03  & 49.52  & 49.13  & 27.86  & 46.11  & 94.09  & 87.88  & 80.92  & \underline{87.64}  & \underline{55.67}  & \underline{59.18}  & \underline{80.66}  & \underline{75.22}  & \underline{65.02}  \\ 
    FedCAD & {63.42}  & \underline{49.17}  & \underline{49.58}  & {49.35}  & \underline{27.94}  & \underline{46.37}  & \underline{94.21}  & \underline{87.94}  & \underline{81.09}  & {87.37}  & {55.59}  & {58.42}  & {79.95}  & {72.81}  & {64.75}  \\ 
    \midrule
    \modelname & \textbf{64.87} & \textbf{50.47} & \textbf{50.71} & \textbf{49.98} & \textbf{28.62} & \textbf{47.05} & \textbf{95.08} & \textbf{88.19} & \textbf{81.73} & \textbf{89.48} & \textbf{56.12} & \textbf{60.53} & \textbf{81.65} & \textbf{75.26} & \textbf{66.39} \\
    (improvement$\uparrow$) & (1.17\%) & (2.64\%) & (2.28\%) & (1.13\%) & (2.43\%) & (1.47\%) & (0.92\%) & (0.28\%) & (0.79\%) & (2.10\%) & (0.81\%) & (2.28\%) & (1.23\%) & (0.05\%) & (2.11\%) \\
    \bottomrule
  \end{tabular}
  \label{tab:detectResults}
  \vspace{-0.3cm}
\end{table*}

\begin{figure}[t]
  \centering
  \includegraphics[width=8.5cm]{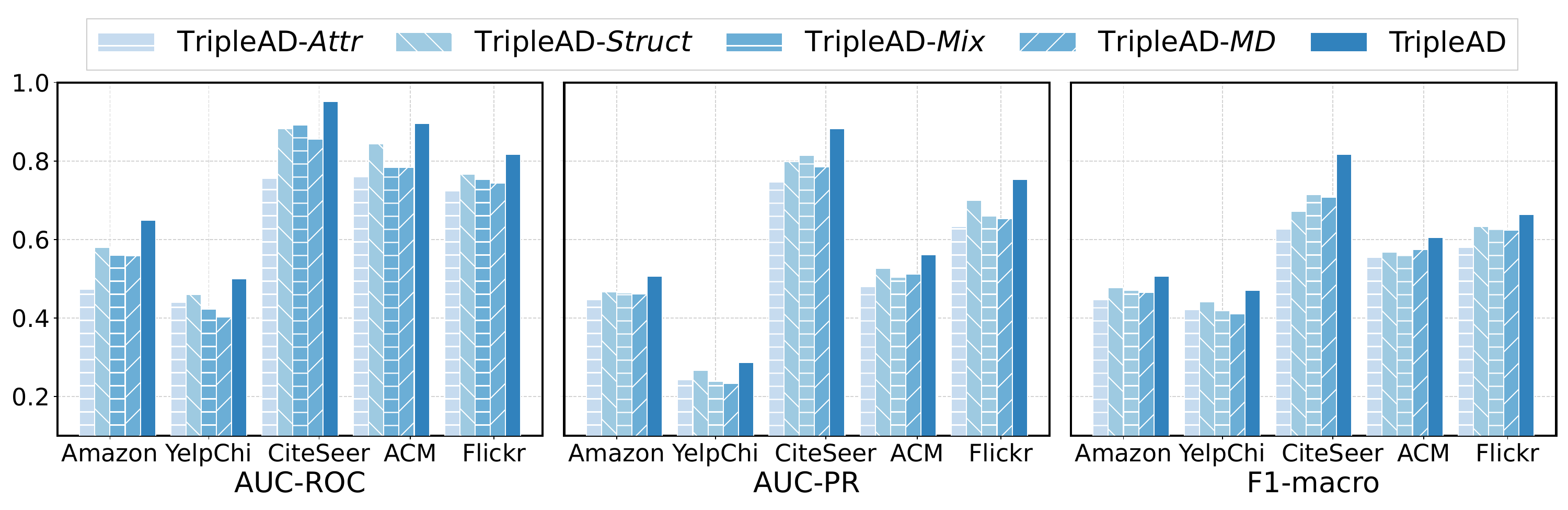}
  \vspace{-0.4cm}  
  \caption{Performance with different components.}
  \label{fig:Strategy}
  \vspace{-4mm}
\end{figure}

\subsection{Ablation Study}

We ablate four important components in \modelname~by removing one of them from the entire model. (1) In \textit{\modelname-Attr}, we remove the multi-scale attribute estimation module; (2) In \textit{\modelname-Struct}, we remove the link-enhanced structure estimation module; (3) In \textit{\modelname-Mix}, we remove the curvature-based mixture estimation module; and (4) In \textit{\modelname-MD}, we remove the mutual distillation scheme. 
As shown in \figurename~\ref{fig:Strategy}, removing any of the three estimation modules (i.e., \textit{\modelname-Attr}, \textit{\modelname-Struct} or \textit{\modelname-Mix})
results in a performance degradation, which indicates that these modules are beneficial for accurately detecting anomalies in graphs.
Besides, the variant \textit{\modelname-MD} does not perform the mutual distillation and ignores the knowledge transfer between the three modules. Lacking the guidance from the structure (resp. attribute) estimation module, the attribute (resp. structure) module generates sub-optimal intermediate representations, leading to decreased performance. 
The same explanation applies to the mixture module.
Notably, \modelname~consistently surpasses all its variants, underlining the efficacy of our designed attribute, structure and mixture  modules in conjunction with the mutual distillation scheme.

\begin{figure}[t]
  \centering
  \includegraphics[width=8.5cm]{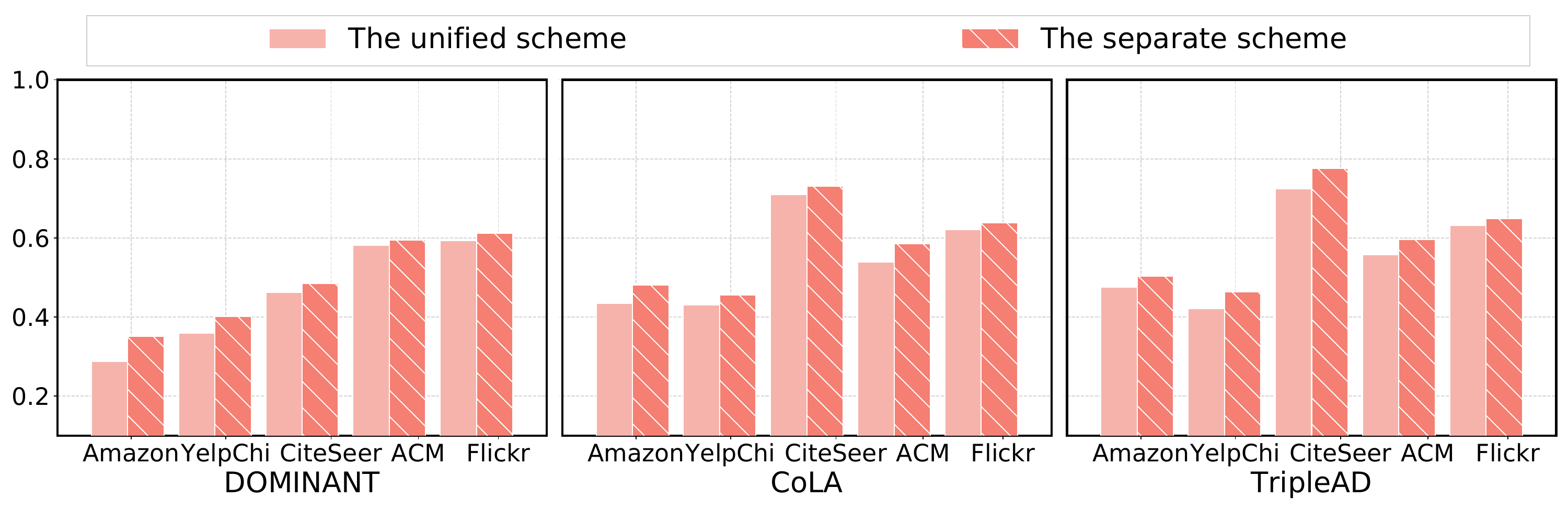}
  \vspace{-0.4cm}  
  \caption{F1-scores using the unified scheme and separate scheme.}
  \label{fig:UniSep}
  \vspace{-0.4cm}
\end{figure}

\subsection{Unified Versus Separate}
\label{sec:separetModel}

To evaluate the effectiveness of our three-channel solution in mitigating the "tug-of-war" problem, we compare the detection performance of a unified model versus separate models. Specifically, we assess both approaches using two representative baselines: the reconstruction-based DOMINANT~\cite{ding2019deep} and the contrast-based CoLA~\cite{liu2021anomaly}, along with the curvature-based mixture estimation module in our model. Under the \textit{unified scheme}, the model is trained and evaluated on both attribute and structural anomalies. Under the \textit{separate scheme}, each model is trained and evaluated first on attribute anomalies and then on structural anomalies. 

\begin{figure}[t]
    \begin{center}
    \subfloat[Anomaly with interference.]    {\label{fig:mutualInterYes}\includegraphics[width=0.23\textwidth]{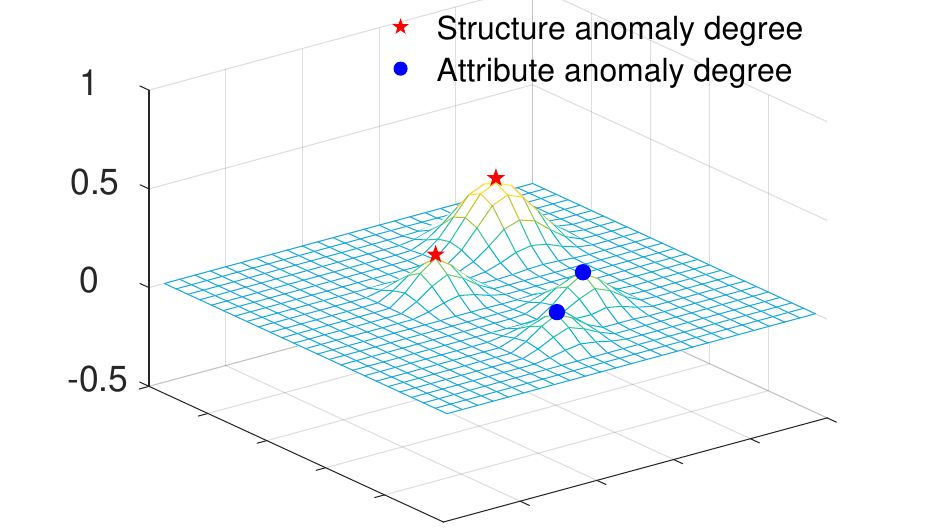}}
    \subfloat[Anomaly without interference.]
    {\label{fig:mutualInterNo}\includegraphics[width=0.23\textwidth]{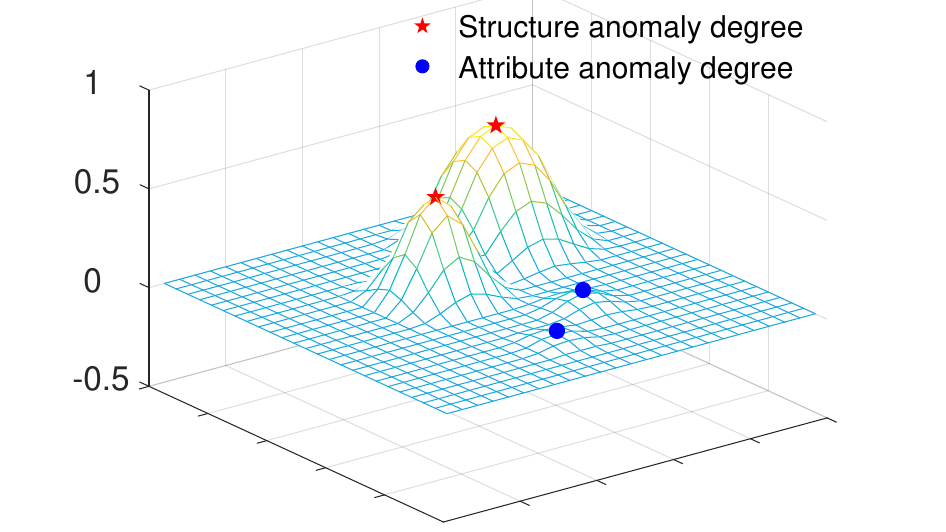}}
    \vspace{-0.1cm}
    \caption{An example of the necessity for mitigating the interference between attribute and structural anomalies. (a) The structure anomaly degree is diluted due to the interference of normal attributes. (b) The structure anomaly degree is more prominent without attribute interference. }
    \label{fig:inteference}
    \end{center}
    \vspace{-0.4cm}
\end{figure}

\figurename~\ref{fig:UniSep} reports the F1 scores of the three methods across five datasets. Notably, 
the \textit{separate scheme} consistently surpasses the \textit{unified scheme}, particularly in datasets with explicitly labeled attribute and structural anomalies (CiteSeer, ACM, and Flickr). In addition, we illustrate the impact of each scheme on anomaly degrees in \figurename~\ref{fig:inteference}. 
These results are computed using DOMINANT on the CiteSeer dataset.
When using the unified model to identify both types of the anomalies, the structural anomaly degree is decreased due to the influences of normal attributes (cf. \figurename~\ref{fig:mutualInterYes}). Conversely, the structural anomaly degree is significantly increased when using the separate model to distinguish structural anomalies (cf. \figurename~\ref{fig:mutualInterNo}). 
We further investigated the performance of models trained on one type of anomaly (e.g., attribute anomalies) when applied to detect another type of anomaly (e.g., structural anomalies). The results indicate that the detection performance in such cases is suboptimal. 
These highlight that the unified scheme is susceptible to mutual interference, leading to degraded performance. In contrast, the separate scheme effectively mitigates the "tug-of-war" dilemma, improving overall performance.

\subsection{Sensitivity Analysis}

\begin{figure}[t]
  \centering
  \includegraphics[width=0.485\textwidth]{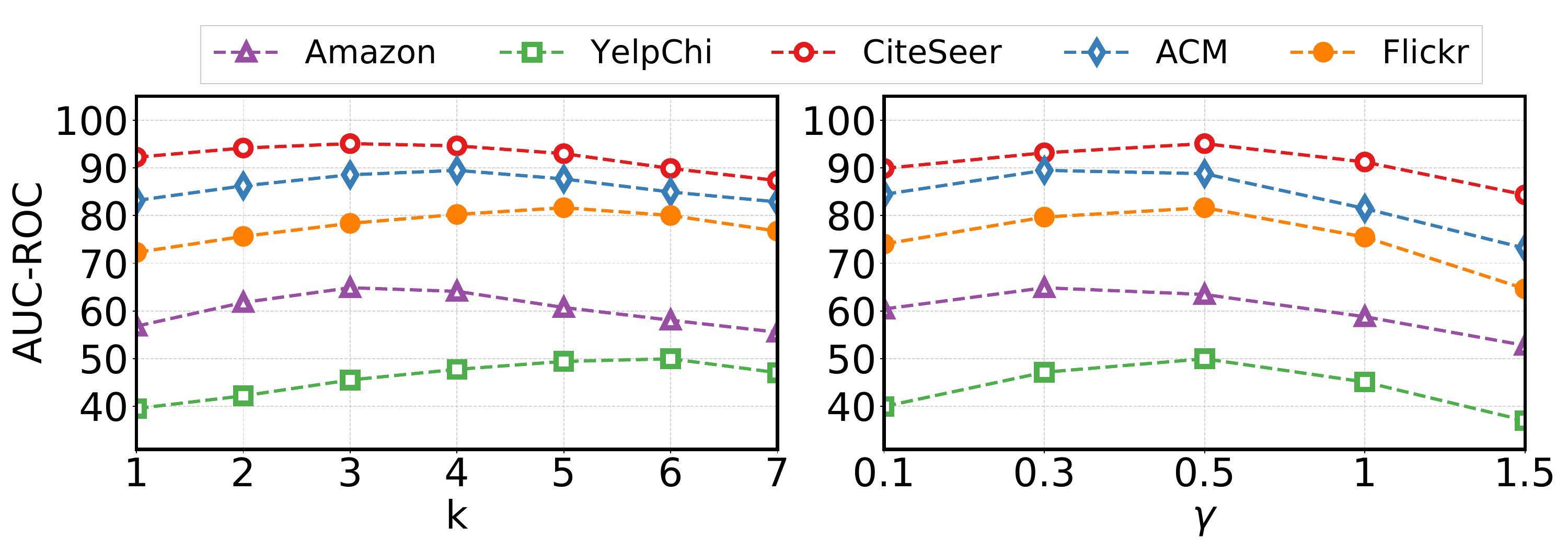}  
  \vspace{-0.7cm}
  \caption{Sensitivity analysis for $k$ and $\gamma$ (AUC).}
  \vspace{-0.4cm}
  \label{fig:lineSensitivity}
\end{figure}

\noindent\textbf{Neighbor Number $k$ and Weighting Coefficient $\gamma$.} 
Here, we alter the selected neighbor number $k$ (cf. Eq.~\eqref{equ:kNN}) and weighting coefficient $\gamma$ (cf. Equation~\eqref{equ:strLoss}) and report their influences on performance for all the datasets in \figurename~\ref{fig:lineSensitivity}. 
Our observations yield the following insights. 
(1) Too large or too small values of $k$ generally undermine performance. 
Since $k$ determines the degree of neighboring nodes connected to the target node, a small $k$ can hinder the model to capture complex information in the graph, and a large $k$ might introduce noises into the learning process.
(2) It can be observed that when the value of $\gamma$ is excessively high, the proportion of the alignment loss significantly surpasses that of the reconstruction loss, leading to a decline in anomaly detection performance.
This is because, while the model aligns the intermediate representations of the two graphs, it loses its ability to conduct effective structural estimation based on the reconstruction loss.

\begin{figure*}[t]
  \centering
  \includegraphics[width=1\textwidth]{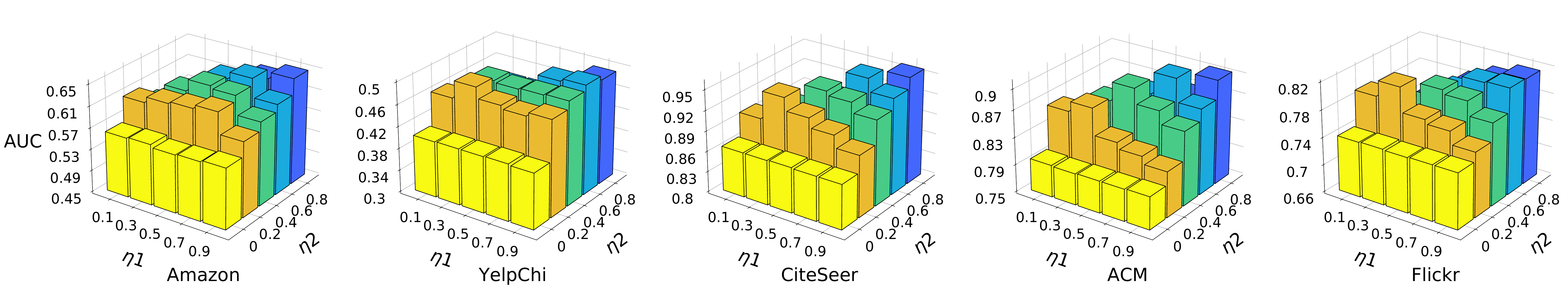}
  \vspace{-7mm}
  \caption{The effects of balancing parameters $\eta_1$ and $\eta_2$ on the five datasets in terms of AUC. }
  \label{fig:3DSensitivity}
  \vspace{-3mm}
\end{figure*}

\textbf{Balancing Parameters $\eta_1$ and $\eta_2$.} 
We here discuss the parameters $\eta_1$ and $\eta_2$ in the loss function (Eqs. \eqref{equ:finalLossAtt},  \eqref{equ:finalLossStr} and \eqref{equ:finalLossMix}).
As shown in Figure~\ref{fig:3DSensitivity}, 
for a given $\eta_1$, 
setting $\eta_2$ to 0 results in the worst performance across all datasets. This observation highlights the importance of our designed distillation loss. While, setting a higher value of $\eta_1$ generally improves the performance. Conversely, an excessively large value of $\eta_2$ is counterproductive -- 
emphasizing too much on the distillation loss will obscure the difference among these three modules and make the corresponding reconstruction an extremely challenging task.

\begin{figure}[htbp]
  \centering
  \includegraphics[width=0.42\textwidth]{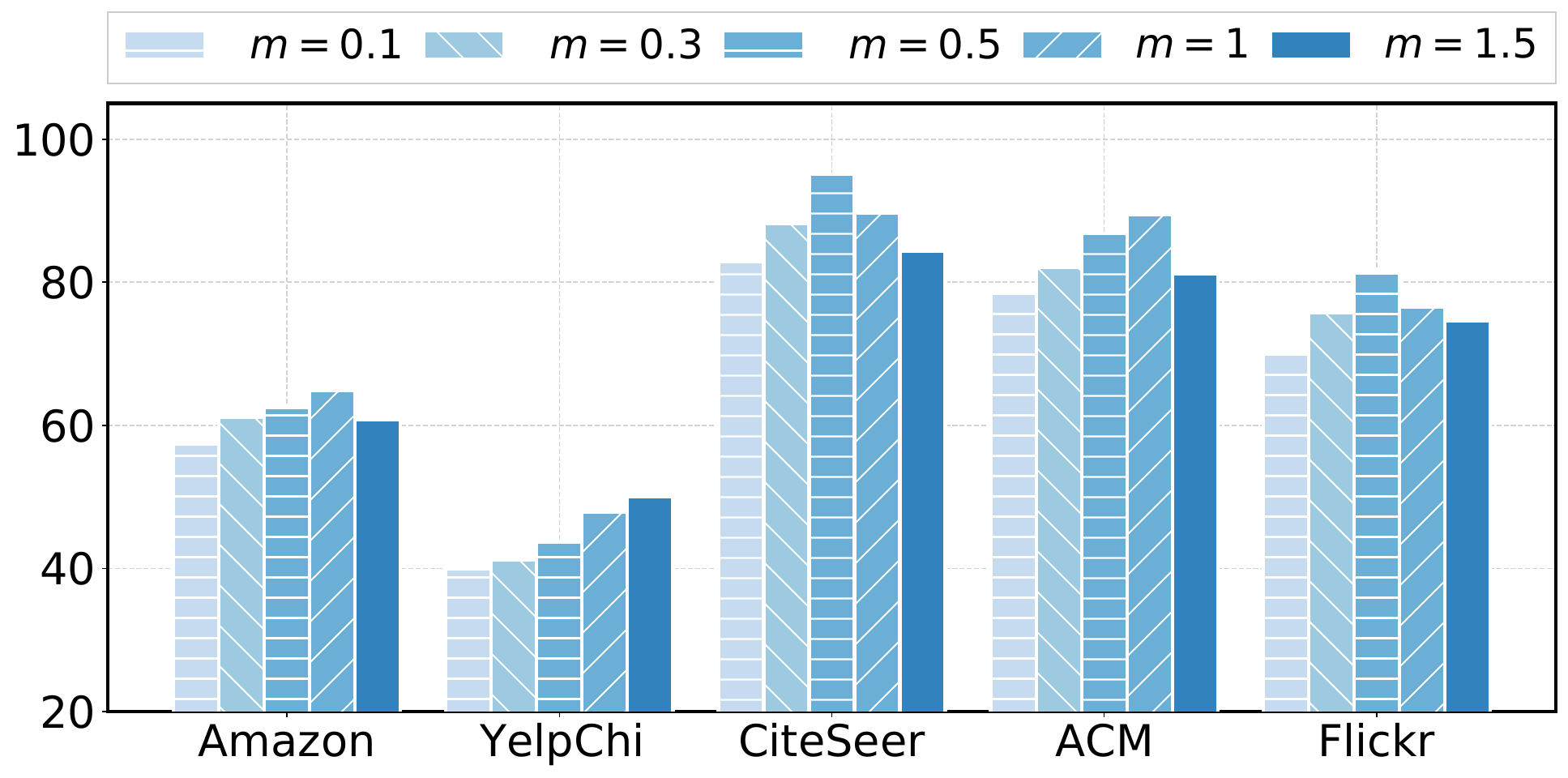}
  \vspace{-0.4cm}
  \caption{Hyper-parameter $m$ w.r.t. AUC.}
  \label{fig:tripletLoss}
  \vspace{-0.4cm}
\end{figure}

\textbf{Triplet Loss Parameter $m$.} 
We experiment with the parameter $m$ in the triplet loss (cf. Eq.~\eqref{equ:distillation}) to determine suitable distances for anchor-positive and anchor-negative pairs within the triplet loss during training. For the five datasets we employed, we maintain consistent parameter settings while varying the values of $m$. We report the values of the AUC-ROC curve on the five datasets in \figurename~\ref{fig:tripletLoss}.
As shown, a lower value of $m$ leads to a poor detection performance. The reason behind is that a lower value will force the teacher and student estimation modules to be more consistent, which diminishes the advantages of our triple-channel strategy.
On the other hand, 
a larger value of $m$ does not necessarily lead to better performance. This means that the appropriate difference between the teacher and student estimation modules is required in order to achieve the optimal detection results.

%% file: 5-discuss.tex
\section{Discussion}

\subsection{Divide and Conquer Strategy}
Our proposed model is inspired by the divide-and-conquer strategy, which has a wide range of applications in natural language processing (NLP)~\cite{chen2023lifelong, chen2023octavius} and computer vision~\cite{wei2020component, tian2021divide}. Nevertheless, this strategy is rarely applied to graph learning. We first discuss the existing applications of the divide-and-conquer strategy and the limitations of this strategy when applied to graph learning. At last, we discuss its potential to graph anomaly detection.  

\textbf{Existing applications of divide and conquer.} 
The divide-and-conquer approach is a foundational strategy in computational algorithms. It aims to divide the input space into several parts and constructs different modules for each part. These modules are then combined into a unified model~\cite{pan2015divide}. This strategy not only simplifies complex problems but also enhances the efficiency and effectiveness of the solutions. It has been widely adopted in various fields, including NLP and visual learning.  
For NLP, researchers primarily integrate the Mixture-of-Experts (MoE) -- which is based on the divide-and-conquer strategy -- for building models such as large language models (LLMs)~\cite{chen2023lifelong, chen2023octavius}. They decompose the language model into several functions, where each ``expert'' learns on a different input space, effectively enhancing the performance of downstream tasks. 
For vision learning, the image is parsed by different components, each of which is trained following a divide-and-conquer learning principle~\cite{tian2021divide, wei2020component}.

\textbf{Challenges for applying it to graph learning}. Despite its success, the divide-and-conquer strategy, including the MOE algorithm, has been rarely explored in the domain of graph learning. The main challenges can be attributed to the following two aspects: (1) The focus of graph learning is to aggregate node attributes along with graph topology for obtaining node representations~\cite{wu2020comprehensive, xia2021graph}; on the other hand, the divide-and-conquer strategy divides the model or data into different parts, conflicting with the idea of graph learning. (2) Dividing graph data into different sub-graphs may lead to information loss, disrupted graph topology, and decreased graph learning performance~\cite{lee2022augmentation}.

\textbf{Divide-and-conquer on graph anomaly detection}. To the best of our knowledge, the divide-and-conquer strategy has not been explored for graph learning. However, we found that it is feasible to apply this strategy to graph anomaly detection: the anomalies can be grouped into three types: attribute, structural, and mixed anomalies~\cite{liu2021anomaly, ma2021comprehensive}. Then the detection of these three types of anomalies can be handled separately. Hence, we introduce the divide-and-conquer strategy into our designed framework and decompose the detection tasks into three distinct channels. Each channel is designed to detect a specific type of anomalies and a mutual distillation module is proposed to encourage knowledge transfer among channels.

\subsection{Relation to GNNs}

We now discuss the relation of our proposed multi-scale feature propagation in the attribute estimation module to the iterative propagation in conventional GNNs.

\begin{figure}[t]
    \begin{center}
    \subfloat[Iteration propagation.]
    {\label{Fig:GNN}\includegraphics[width=0.161\textwidth]{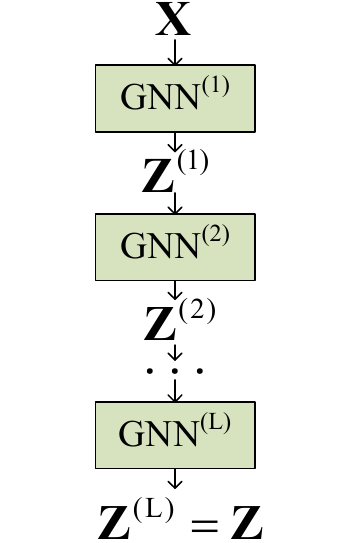}}
    \subfloat[Multi-scale feature propagation.]
    {\label{Fig:Multi-scale}\includegraphics[width=0.295\textwidth]{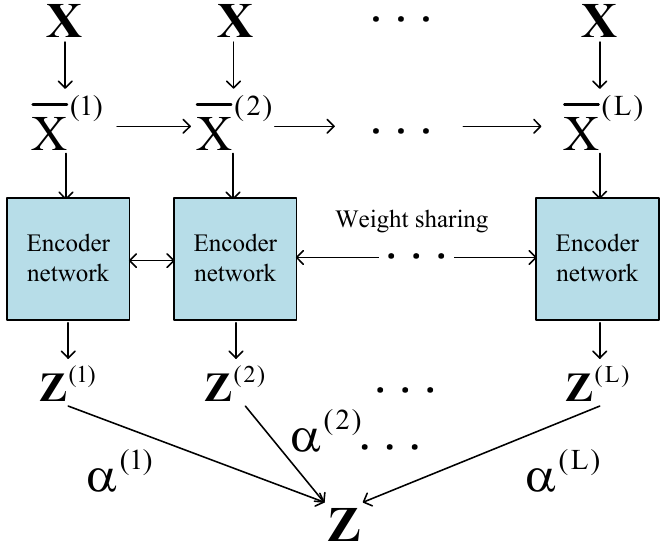}}
    \vspace{-0.1cm}
    \caption{Relation of the iterative propagation in GNNs and the multi-scale feature propagation in the proposed module.}
    \label{Fig:GNN_Multi-scale}
    \end{center}
    \vspace{-0.5cm}
\end{figure}

\textbf{Iterative Propagation in Conventional GNNs.} As illustrated in Figure \ref{Fig:GNN}, given the adjacency matrix $\mathbf{A}$ and corresponding feature matrix $\mathbf{X}$, the conventional GNNs compute intermediate representations by iteratively performing propagation/aggregation and transformation:
\begin{equation} \label{equ:conventionalGNN}
    \begin{aligned}
        \mathbf{Z}^{(l)} &= \text{GNN}^{(l)}(\mathbf{A}, \mathbf{Z}^{(l-1)}) \\
        &= \sigma(\bar{\mathbf{A}}\mathbf{Z}^{(l-1)}\mathbf{W}^{(l)}),
    \end{aligned}
\end{equation}
where $l \in [ 1, L ]$, $\mathbf{Z}^{(0)} = \mathbf{X}$, $\mathbf{Z} = \mathbf{Z}^{L}$ represents the final learned intermediate representation from the GNNs, $\sigma(\cdot)$ is a non-linear activation function, such as $\text{ReLU}$, and $\mathbf{W}^{(l)}$ denotes the learnable matrix of each $\text{GNN}^{(l)}$ layer.

\textbf{Multi-scale Feature Propagation.} 
From the Figure \ref{Fig:Multi-scale}, it can be observed that the multi-scale feature propagation performs propagation at different scales, generating $L$ propagation matrices $\{\bar{\mathbf{X}}^{(l)}\}_{l=1}^{L}$. Each matrix $\mathbf{X}^{(l)}$ aggregates features from a one-hop neighbors:
\begin{equation}\label{equ:multiScale}
    \bar{\mathbf{X}}^{(l)} = (1-\alpha ) \bar{\mathbf{A}} \mathbf{X}^{(l-1)} + \alpha \mathbf{X},
\end{equation}
where $l \in [ 1, L ]$, $\mathbf{X}^{0} = \mathbf{X}$, $\alpha \in$ (0,1] is the restart probability, and $\bar{\mathbf{A}}$ is the normalized adjacency matrix. Then, the propagation matrices $\{\bar{\mathbf{X}}^{(l)}\}_{l=1}^{L}$ are encoded by a parameter-shared encoder to obtain $L$ intermediate representations:
\begin{equation}\label{equ:encoder}
    \mathbf{Z}^{(l)} = \sigma(\bar{\mathbf{X}}^{(l)}\mathbf{W}),
\end{equation}
where $\mathbf{W}$ is the learnable matrix whose parameters are shared with the encoder. 
Then, the intermediate representations $\{\mathbf{Z}^{(l)}\}_{l=1}^{L}$ are combined to generate the final representation $\mathbf{Z}$ by sum operation weighted by the attention score $\boldsymbol{\alpha}^{(l)}$:
\begin{equation}\label{equ:attentionMix}
    \mathbf{Z} = \sum \boldsymbol{\alpha}^{(l)}\mathbf{Z}^{(l)}.
\end{equation}


Compared to the iterative propagation in traditional GNNs, the multi-scale feature propagation we designed broadens the scope of feature propagation while decreasing its depth. 
While, unlike traditional GNNs where multiple iterations are required, our model performs each step only once and therefore, reducing the parameters introduced by multiple GNN layers. It not only increases the efficiency of model training, but also avoids the over-smoothing issue caused by deep GNNs.


\subsection{Complexity Analysis}

The primary computational overheads of \modelname~lie in three aspects: the multi-scale attribute, the link-enhanced structure, and the curvature-based mixture estimation modules. 
%
The time complexity of the propagation in the attribute module is $\mathcal{O}(NLd)$, where $N$ is the number of nodes, $L$ is the number of propagation scales, and $d$ is the representation dimension. During training of attribute estimation, the combination of $L$ representations through the attention mechanism incurs a complexity of $\mathcal{O}(L^2hd)$, where $h$ is the dimension of $\mathbf{Z}^{(l)}$. 
In the structure module, the complexities of the pre-processing for the masked graph and the enhanced graph are $\mathcal{O}(NTd)$ and $\mathcal{O}(NkTd)$, respectively, where $k$ is the number of neighbors in the $k$NN graph, $T$ is the number of iterations. 
In the training of structure estimation, the time complexity per epoch is $\mathcal{O}(Nhd)$. 
In the mixture module, the predominant computational cost arises from defining the probability distribution of nodes and computing the attribute-mixed curvatures, and their respective time complexities are $\mathcal{O}(N^2d)$ and $\mathcal{O}(|\mathcal{E}|N^2)$, respectively, where $|\mathcal{E}|$ is the number of edges.

%% file: 6-relate.tex
\section{Relate Work}

\subsection{Graph Anomaly Detection}

Initial efforts in graph anomaly detection utilized shallow techniques 
to identify anomalous patterns. These approaches include ego-graph~\cite{perozzi2016scalable}, residual analysis~\cite{li2017radar}, CUR decomposition \cite{peng2018anomalous}, and clustering methods~\cite{liu2017accelerated}. However, these methods suffer from issues such as network sparsity and data nonlinearity, which limit their ability to capture complex interrelations among nodes in graphs. 
Due to the advancements of deep graph learning~\cite{wu2024high, mo2023multiplex, yu2019self}, many deep models have been proposed and achieved considerable success. 
Some of these models leverage labeled data as supervision signals to identify anomalies~\cite{gao2023addressing, xiao2024Motif, gao2023alleviating}. 
However, due to the difficulty of collecting anomalous labels, a greater emphasis has been placed on unsupervised approaches,
which fall into two groups: reconstruction-based and contrast-based methods.  

The reconstruction-based approaches focus on reconstructing node attributes and structure, and utilizing the reconstruction errors to identify anomalous nodes~\cite{ding2019deep, he2024ada}. 
For example, DOMINANT~\cite{ding2019deep} employs an autoencoder framework to learn node representations through the reconstruction of both attribute and structural information. 
Based on this framework, graph convolutional networks (GCNs) and deep attention mechanism have been introduced for graph anomaly detection~\cite{huang2021hybrid, shao2023learning}. 
Further, ComGA~\cite{luo2022comga} exploits graph community structure and AnomMAN~\cite{chen2023anomman} learn diverse data distributions from multiple perspectives for reconstruction.
Moreover, AnomalyDAE~\cite{fan2020anomalydae} learns two separate representations for attribute and structure and 
combine the learned representations to enhance reconstruction.

The contrast-based methods, on the other hand, learn node-level and subgraph-level representations and then adopt level agreements to detect anomalies~\cite{duan2023arise, hu2023samcl}. 
For example, 
CoLA~\cite{liu2021anomaly} introduces contrastive learning to detect anomalies by contrasting nodes with their neighborhoods. 
ANEMONE~\cite{jin2021anemone} performs patch-level and context-level contrastive learning, facilitating anomaly detection through statistical analyses of contrastive scores.
Sub-CR~\cite{zhang2022reconstruction} conducts intra-view and inter-view contrastive learning and then integrates both modules based on attribute reconstruction. 
GRADATE~\cite{duan2023graph} constructs a multi-view contrastive network that includes node-subgraph, node-node, and subgraph-subgraph comparisons, combining diverse anomaly information to compute node anomaly scores.

\begin{table}[htbp]
  \fontsize{7.2}{9}\selectfont
  \caption{Comparison of related works.}
  \label{table:related}
  \vspace{-0.3cm}
  \centering
  \begin{threeparttable} 
  \newcolumntype{M}[1]{>{\centering\arraybackslash}m{#1}}
  \begin{tabular}{M{0.25cm}<{\centering}|p{1.90cm}<{\centering}|p{0.72cm}<{\centering} p{0.72cm}<{\centering} p{0.56cm}<{\centering} p{0.70cm}<{\centering} p{1.09cm}<{\centering}}
     \toprule
     &Methods &Attribute\tnote{1} &Structure\tnote{1} &Mixed\tnote{1} &Separate\tnote{2}& Interference\tnote{3}\\
    \midrule
    \multirow{5}{*}{\rotatebox[origin=c]{90}{Shallow}}  
    & AMEN~\cite{perozzi2016scalable}      & \checkmark  & \checkmark  &    &   & n/a  \\
    & Radar~\cite{li2017radar}            & \checkmark  & \checkmark  &    &   & n/a  \\
    & ANOMALOUS~\cite{peng2018anomalous}  & \checkmark  & \checkmark  &    &   & n/a  \\
    & Mixedad~\cite{zhu2020mixedad}       & \checkmark  & \checkmark  & \checkmark  &   & n/a \\
    & ALAD~\cite{liu2017accelerated}      &         &   & \checkmark  &   &  n/a  \\
    \midrule
    \multirow{7}{*}{\rotatebox[origin=c]{90}{Reconstruction}} 
    & DOMINANT~\cite{ding2019deep}        & \checkmark  & \checkmark  &   &   &   \\
    & AnomalyDAE~\cite{fan2020anomalydae} & \checkmark  & \checkmark  &   &\checkmark   &   \\
    & HO-GAT~\cite{huang2021hybrid}        & \checkmark  & \checkmark  &   &   &   \\
    & GDAE~\cite{shao2023learning}         & \checkmark  & \checkmark  &   &   &   \\
    & ComGA~\cite{luo2022comga}           & \checkmark  & \checkmark  & \checkmark &   &   \\
    & AnomMAN~\cite{chen2023anomman}      & \checkmark  & \checkmark  &   &   &   \\
    & ADA-GAD~\cite{he2024ada}            & \checkmark  & \checkmark  &   &   &   \\    
    & GAD-NR~\cite{roy2024gad}    & \checkmark  & \checkmark  &   &   &   \\
    \midrule
    \multirow{4}{*}{\rotatebox[origin=c]{90}{Contrast}}  
    & CoLA~\cite{liu2021anomaly}          & \checkmark  & \checkmark  &   &   &   \\
    & ANEMONE~\cite{jin2021anemone}        & \checkmark  & \checkmark  &   &   &   \\
    & Sub-CR~\cite{zhang2022reconstruction}& \checkmark  & \checkmark  &   &   &   \\
    & GRADATE~\cite{duan2023graph}         & \checkmark  & \checkmark  &   &   &   \\
    & FedCAD~\cite{kong2024federated} & \checkmark  & \checkmark  &   &   & \\
    \midrule
    & \modelname~(ours) & \checkmark & \checkmark & \checkmark & \checkmark & \checkmark  \\
    \bottomrule
  \end{tabular}
  \begin{tablenotes}
  \footnotesize 
     \item[1] \textit{Attribute}, \textit{Structure} and \textit{Mixed} denote that the model explicitly focuses on detecting attribute, structure and mixed anomalies, respectively.     
    \item[2] \textit{Separate} means that different modules are devised for distinct representations.
    \item[3] \textit{Interference} represents that the detection models can alleviate the interference between attributes and structure.
  \end{tablenotes}
  \end{threeparttable} 
  \vspace{-0.3cm}
\end{table}

The differences between existing unsupervised anomaly detection methods and our \modelname~are summarized in Table~\ref{table:related}.
Existing deep learning models do not adequately resolve the interference (or ``tug-of-war'') between attribute and structural representation learning. 
Note that AnomalyDAE~\cite{fan2020anomalydae} tries to learn attribute and structural representations in different channels. However, they concatenate the distinct representations to simultaneously detect both types of anomalies, which cannot avoid the interference between attribute and structure anomaly detection.

\subsection{Graph Imputation}

Our model masks a part of attributes or edges and then estimates the masked values, aligning closely with graph imputation. There are two main graph imputation tasks: 
attribute imputation and structural imputation.
Attribute imputation focuses on recovering missing attributes within graphs~\cite{ding2022data}. In this aspect, 
SAT~\cite{chen2020learning} introduces a feature-structure distribution matching mechanism for attribute imputation.
HGNN-AC~\cite{jin2021heterogeneous} employs topological embeddings to improve attribute completion.
ITR~\cite{tu2022initializing} utilizes structural information for initial imputation, subsequently refining the imputed latent variables using existing attribute and structural information. 
Additionally, GNN-based autoencoder and Gaussian mixture model have been introduced to fill missing attributes~\cite{spinelli2020missing, taguchi2021graph, gao2023handling}.

Structural imputation (also called link prediction) aims to infer missing links within a graph~\cite{tan2023s2gae,wang2020inductive}. The mainstream way is to learn representations for the nodes at either end of a potential link, and then combining these representations to compute link existence probability~\cite{yun2021neo}. 
Based on this idea, NBFNet~\cite{zhu2021neural} and LLP~\cite{guo2023linkless} utilize a GNN-based encoder to learn node representations and then employ a decoder to predict link existence. 
Further, counterfactual learning, self-supervised learning and reinforcement learning are introduced to improve link prediction performance~\cite{zhao2022learning, liu2024self, wang2019learning}.

%% file: 7-conclusion.tex
\section{Conclusion}
In this paper, we presented a novel mutual distillation-based Triple-channel graph Anomaly Detection framework (\modelname). It can effectively relieve the interference between attributes and structure by learning three different but collaborated representations for detecting attribute, structural and mixed anomalies, respectively.
In terms of the attribute channel, we designed a multi-scale attribute estimation module, which explores multiple augmented views based on different feature propagation scales to alleviate the over-smoothing issue in traditional GNNs and advocate attribute anomaly detection. 
In terms of the structure channel, we devised a structure estimation module that generates a link-enhanced graph to promote information sharing to the isolated nodes and boost structure anomaly detection.
In terms of the mixture channel, we presented a curvature-based mixture estimation module, which introduces a new attribute-mixed curvature to reflect the attribute and structure information for mixed anomaly detection.
Besides, we proposed a mutual distillation strategy that employs a teacher-student distillation to exchange knowledge from the three channels and refine learned node representations.
Extensive experiments demonstrated that the proposed \M~can alleviate the mutual interference problem and improve anomaly detection performance.

%% file: 0-TripleAD.bbl
\begin{thebibliography}{10}
\providecommand{\url}[1]{#1}
\csname url@samestyle\endcsname
\providecommand{\newblock}{\relax}
\providecommand{\bibinfo}[2]{#2}
\providecommand{\BIBentrySTDinterwordspacing}{\spaceskip=0pt\relax}
\providecommand{\BIBentryALTinterwordstretchfactor}{4}
\providecommand{\BIBentryALTinterwordspacing}{\spaceskip=\fontdimen2\font plus
\BIBentryALTinterwordstretchfactor\fontdimen3\font minus \fontdimen4\font\relax}
\providecommand{\BIBforeignlanguage}[2]{{%
\expandafter\ifx\csname l@#1\endcsname\relax
\typeout{** WARNING: IEEEtran.bst: No hyphenation pattern has been}%
\typeout{** loaded for the language `#1'. Using the pattern for}%
\typeout{** the default language instead.}%
\else
\language=\csname l@#1\endcsname
\fi
#2}}
\providecommand{\BIBdecl}{\relax}
\BIBdecl

\bibitem{liu2024generalized}
Y.~Liu, D.~Yang, Y.~Wang, J.~Liu, J.~Liu, A.~Boukerche, P.~Sun, and L.~Song, ``Generalized video anomaly event detection: Systematic taxonomy and comparison of deep models,'' \emph{ACM Computing Surveys}, vol.~56, no.~7, pp. 1--38, 2024.

\bibitem{zhang2024self}
K.~Zhang, Q.~Wen, C.~Zhang, R.~Cai, M.~Jin, Y.~Liu, J.~Y. Zhang, Y.~Liang, G.~Pang, D.~Song \emph{et~al.}, ``Self-supervised learning for time series analysis: Taxonomy, progress, and prospects,'' \emph{IEEE Transactions on Pattern Analysis and Machine Intelligence}, vol.~46, no.~10, pp. 6775--6794, 2024.

\bibitem{ma2023fighting}
J.~Ma, F.~Li, R.~Zhang, Z.~Xu, D.~Cheng, Y.~Ouyang, R.~Zhao, J.~Zheng, Y.~Zheng, and C.~Jiang, ``Fighting against organized fraudsters using risk diffusion-based parallel graph neural network.'' in \emph{International Joint Conference on Artificial Intelligence}, 2023, pp. 6138--6146.

\bibitem{wu2020comprehensive}
Z.~Wu, S.~Pan, F.~Chen, G.~Long, C.~Zhang, and S.~Y. Philip, ``A comprehensive survey on graph neural networks,'' \emph{IEEE Transactions on Neural Networks and Learning Systems}, vol.~32, no.~1, pp. 4--24, 2020.

\bibitem{luo2022comga}
X.~Luo, J.~Wu, A.~Beheshti, J.~Yang, X.~Zhang, Y.~Wang, and S.~Xue, ``Comga: Community-aware attributed graph anomaly detection,'' in \emph{ACM International Conference on Web Search and Data Mining}, 2022, pp. 657--665.

\bibitem{branco2020interleaved}
B.~Branco, P.~Abreu, A.~S. Gomes, M.~S. Almeida, J.~T. Ascens{\~a}o, and P.~Bizarro, ``Interleaved sequence rnns for fraud detection,'' in \emph{Proceedings of the ACM SIGKDD Conference on Knowledge Discovery and Data Mining}, 2020, pp. 3101--3109.

\bibitem{nguyen2020fang}
V.-H. Nguyen, K.~Sugiyama, P.~Nakov, and M.-Y. Kan, ``Fang: Leveraging social context for fake news detection using graph representation,'' in \emph{ACM International Conference on Information \& Knowledge Management}, 2020, pp. 1165--1174.

\bibitem{yu2018netwalk}
W.~Yu, W.~Cheng, C.~C. Aggarwal, K.~Zhang, H.~Chen, and W.~Wang, ``Netwalk: A flexible deep embedding approach for anomaly detection in dynamic networks,'' in \emph{Proceedings of the ACM SIGKDD Conference on Knowledge Discovery and Data Mining}, 2018, pp. 2672--2681.

\bibitem{liu2022benchmarking}
K.~Liu, Y.~Dou, Y.~Zhao, X.~Ding, X.~Hu, R.~Zhang, K.~Ding, C.~Chen, H.~Peng, K.~Shu \emph{et~al.}, ``Bond: Benchmarking unsupervised outlier node detection on static attributed graphs,'' in \emph{Advances in Neural Information Processing Systems}, vol.~35, 2022, pp. 27\,021--27\,035.

\bibitem{xiao2024counterfactual}
C.~Xiao, S.~Pang, X.~Xu, X.~Li, G.~Trajcevski, and F.~Zhou, ``Counterfactual data augmentation with denoising diffusion for graph anomaly detection,'' \emph{IEEE Transactions on Computational Social Systems}, vol.~11, no.~6, pp. 7555--7567, 2024.

\bibitem{li2017radar}
J.~Li, H.~Dani, X.~Hu, and H.~Liu, ``Radar: Residual analysis for anomaly detection in attributed networks,'' in \emph{International Joint Conference on Artificial Intelligence}, 2017, pp. 2152--2158.

\bibitem{zhu2020mixedad}
M.~Zhu and H.~Zhu, ``Mixedad: A scalable algorithm for detecting mixed anomalies in attributed graphs,'' in \emph{AAAI Conference on Artificial Intelligence}, 2020, pp. 1274--1281.

\bibitem{liu2021anomaly}
Y.~Liu, Z.~Li, S.~Pan, C.~Gong, C.~Zhou, and G.~Karypis, ``Anomaly detection on attributed networks via contrastive self-supervised learning,'' \emph{IEEE Transactions on Neural Networks and Learning Systems}, vol.~33, no.~6, pp. 2378--2392, 2021.

\bibitem{ma2021comprehensive}
X.~Ma, J.~Wu, S.~Xue, J.~Yang, C.~Zhou, Q.~Z. Sheng, H.~Xiong, and L.~Akoglu, ``A comprehensive survey on graph anomaly detection with deep learning,'' \emph{IEEE Transactions on Knowledge and Data Engineering}, vol.~35, no.~12, pp. 12\,012--12\,038, 2021.

\bibitem{ding2019deep}
K.~Ding, J.~Li, R.~Bhanushali, and H.~Liu, ``Deep anomaly detection on attributed networks,'' in \emph{SIAM International Conference on Data Mining}, 2019, pp. 594--602.

\bibitem{huang2021hybrid}
L.~Huang, Y.~Zhu, Y.~Gao, T.~Liu, C.~Chang, C.~Liu, Y.~Tang, and C.-D. Wang, ``Hybrid-order anomaly detection on attributed networks,'' \emph{IEEE Transactions on Knowledge and Data Engineering}, vol.~35, no.~12, pp. 12\,249--12\,263, 2021.

\bibitem{peng2020deep}
Z.~Peng, M.~Luo, J.~Li, L.~Xue, and Q.~Zheng, ``A deep multi-view framework for anomaly detection on attributed networks,'' \emph{IEEE Transactions on Knowledge and Data Engineering}, vol.~34, no.~6, pp. 2539--2552, 2020.

\bibitem{zheng2021generative}
Y.~Zheng, M.~Jin, Y.~Liu, L.~Chi, K.~T. Phan, and Y.-P.~P. Chen, ``Generative and contrastive self-supervised learning for graph anomaly detection,'' \emph{IEEE Transactions on Knowledge and Data Engineering}, vol.~35, no.~12, pp. 12\,220 -- 12\,233, 2021.

\bibitem{zhang2022reconstruction}
J.~Zhang, S.~Wang, and S.~Chen, ``Reconstruction enhanced multi-view contrastive learning for anomaly detection on attributed networks,'' in \emph{International Joint Conference on Artificial Intelligence}, 2022, pp. 2376--2382.

\bibitem{duan2023graph}
J.~Duan, S.~Wang, P.~Zhang, E.~Zhu, J.~Hu, H.~Jin, Y.~Liu, and Z.~Dong, ``Graph anomaly detection via multi-scale contrastive learning networks with augmented view,'' in \emph{AAAI Conference on Artificial Intelligence}, 2023, pp. 7459--7467.

\bibitem{hadsell2020embracing}
R.~Hadsell, D.~Rao, A.~A. Rusu, and R.~Pascanu, ``Embracing change: Continual learning in deep neural networks,'' \emph{Trends in cognitive sciences}, vol.~24, no.~12, pp. 1028--1040, 2020.

\bibitem{chen2023octavius}
Z.~Chen, Z.~Wang, Z.~Wang, H.~Liu, Z.~Yin, S.~Liu, L.~Sheng, W.~Ouyang, Y.~Qiao, and J.~Shao, ``Octavius: Mitigating task interference in mllms via moe,'' in \emph{International Conference on Learning Representations}, 2024.

\bibitem{yang2022graph}
L.~Yang, W.~Zhou, W.~Peng, B.~Niu, J.~Gu, C.~Wang, X.~Cao, and D.~He, ``Graph neural networks beyond compromise between attribute and topology,'' in \emph{Proceedings of the ACM Web Conference}, 2022, pp. 1127--1135.

\bibitem{wang2020gcn}
X.~Wang, M.~Zhu, D.~Bo, P.~Cui, C.~Shi, and J.~Pei, ``Am-gcn: Adaptive multi-channel graph convolutional networks,'' in \emph{Proceedings of the ACM SIGKDD Conference on Knowledge Discovery and Data Mining}, 2020, pp. 1243--1253.

\bibitem{tian2021divide}
Y.~Tian, O.~J. Henaff, and A.~van~den Oord, ``Divide and contrast: Self-supervised learning from uncurated data,'' in \emph{IEEE/CVF International Conference on Computer Vision}, 2021, pp. 10\,063--10\,074.

\bibitem{wei2020component}
P.~Wei, Z.~Xie, H.~Lu, Z.~Zhan, Q.~Ye, W.~Zuo, and L.~Lin, ``Component divide-and-conquer for real-world image super-resolution,'' in \emph{European Conference on Computer Vision}, 2020, pp. 101--117.

\bibitem{gidiotis2020divide}
A.~Gidiotis and G.~Tsoumakas, ``A divide-and-conquer approach to the summarization of long documents,'' \emph{IEEE/ACM Transactions on Audio, Speech, and Language Processing}, vol.~28, pp. 3029--3040, 2020.

\bibitem{ollivier2009ricci}
Y.~Ollivier, ``Ricci curvature of markov chains on metric spaces,'' \emph{Journal of Functional Analysis}, vol. 256, no.~3, pp. 810--864, 2009.

\bibitem{ye2019curvature}
Z.~Ye, K.~S. Liu, T.~Ma, J.~Gao, and C.~Chen, ``Curvature graph network,'' in \emph{International Conference on Learning Representations}, 2020.

\bibitem{guo2021learning}
X.~Guo, Q.~Tian, W.~Zhang, W.~Wang, and P.~Jiao, ``Learning stochastic equivalence based on discrete ricci curvature,'' in \emph{International Joint Conference on Artificial Intelligence}, 2021, pp. 1456--1462.

\bibitem{li2022curvature}
H.~Li, J.~Cao, J.~Zhu, Y.~Liu, Q.~Zhu, and G.~Wu, ``Curvature graph neural network,'' \emph{Information Sciences}, vol. 592, pp. 50--66, 2022.

\bibitem{chen2020measuring}
D.~Chen, Y.~Lin, W.~Li, P.~Li, J.~Zhou, and X.~Sun, ``Measuring and relieving the over-smoothing problem for graph neural networks from the topological view,'' in \emph{AAAI Conference on Artificial Intelligence}, vol.~34, no.~04, 2020, pp. 3438--3445.

\bibitem{ding2023eliciting}
K.~Ding, Y.~Wang, Y.~Yang, and H.~Liu, ``Eliciting structural and semantic global knowledge in unsupervised graph contrastive learning,'' in \emph{AAAI Conference on Artificial Intelligence}, 2023, pp. 7378--7386.

\bibitem{xie2022self}
Y.~Xie, Z.~Xu, and S.~Ji, ``Self-supervised representation learning via latent graph prediction,'' in \emph{International Conferenceon Machine Learning}, 2022, pp. 24\,460--24\,477.

\bibitem{xiao2023counterfactual}
C.~Xiao, X.~Xu, Y.~Lei, K.~Zhang, S.~Liu, and F.~Zhou, ``Counterfactual graph learning for anomaly detection on attributed networks,'' \emph{IEEE Transactions on Knowledge and Data Engineering}, vol.~35, no.~10, pp. 10\,540--10\,553, 2023.

\bibitem{liu2023learning}
Y.~Liu, K.~Ding, J.~Wang, V.~Lee, H.~Liu, and S.~Pan, ``Learning strong graph neural networks with weak information,'' in \emph{Proceedings of the ACM SIGKDD Conference on Knowledge Discovery and Data Mining}, 2023, pp. 1559--1571.

\bibitem{kipf2016semi}
T.~N. Kipf and M.~Welling, ``Semi-supervised classification with graph convolutional networks,'' in \emph{International Conference on Learning Representations}, 2017.

\bibitem{gou2021knowledge}
J.~Gou, B.~Yu, S.~J. Maybank, and D.~Tao, ``Knowledge distillation: A survey,'' \emph{International Journal of Computer Vision}, vol. 129, no.~6, pp. 1789--1819, 2021.

\bibitem{tian2025knowledge}
Y.~Tian, S.~Pei, X.~Zhang, C.~Zhang, and N.~V. Chawla, ``Knowledge distillation on graphs: A survey,'' \emph{ACM Computing Surveys}, vol.~57, no.~8, pp. 1--16, 2025.

\bibitem{oki2020triplet}
H.~Oki, M.~Abe, J.~Miyao, and T.~Kurita, ``Triplet loss for knowledge distillation,'' in \emph{International Joint Conference on Neural Networks}, 2020, pp. 1--7.

\bibitem{boutros2022self}
F.~Boutros, N.~Damer, F.~Kirchbuchner, and A.~Kuijper, ``Self-restrained triplet loss for accurate masked face recognition,'' \emph{Pattern Recognition}, vol. 124, p. 108473, 2022.

\bibitem{tong2011non}
H.~Tong and C.-Y. Lin, ``Non-negative residual matrix factorization with application to graph anomaly detection,'' in \emph{SIAM International Conference on Data Mining}, 2011, pp. 143--153.

\bibitem{tang2022rethinking}
J.~Tang, J.~Li, Z.~Gao, and J.~Li, ``Rethinking graph neural networks for anomaly detection,'' in \emph{International Conferenceon Machine Learning}, 2022, pp. 21\,076--21\,089.

\bibitem{ding2021inductive}
K.~Ding, J.~Li, N.~Agarwal, and H.~Liu, ``Inductive anomaly detection on attributed networks,'' in \emph{International Joint Conference on Artificial Intelligence}, 2021, pp. 1288--1294.

\bibitem{perozzi2016scalable}
B.~Perozzi and L.~Akoglu, ``Scalable anomaly ranking of attributed neighborhoods,'' in \emph{SIAM International Conference on Data Mining}, 2016, pp. 207--215.

\bibitem{peng2018anomalous}
Z.~Peng, M.~Luo, J.~Li, H.~Liu, Q.~Zheng \emph{et~al.}, ``Anomalous: A joint modeling approach for anomaly detection on attributed networks,'' in \emph{International Joint Conference on Artificial Intelligence}, 2018, pp. 3513--3519.

\bibitem{fan2020anomalydae}
H.~Fan, F.~Zhang, and Z.~Li, ``Anomalydae: Dual autoencoder for anomaly detection on attributed networks,'' in \emph{IEEE International Conference on Acoustics, Speech and Signal Processing}, 2020, pp. 5685--5689.

\bibitem{he2024ada}
J.~He, Q.~Xu, Y.~Jiang, Z.~Wang, and Q.~Huang, ``Ada-gad: Anomaly-denoised autoencoders for graph anomaly detection,'' in \emph{AAAI Conference on Artificial Intelligence}, vol.~38, no.~8, 2024, pp. 8481--8489.

\bibitem{roy2024gad}
A.~Roy, J.~Shu, J.~Li, C.~Yang, O.~Elshocht, J.~Smeets, and P.~Li, ``Gad-nr: Graph anomaly detection via neighborhood reconstruction,'' in \emph{ACM International Conference on Web Search and Data Mining}, 2024, pp. 576--585.

\bibitem{kong2024federated}
X.~Kong, W.~Zhang, H.~Wang, M.~Hou, X.~Chen, X.~Yan, and S.~K. Das, ``Federated graph anomaly detection via contrastive self-supervised learning,'' \emph{IEEE Transactions on Neural Networks and Learning Systems}, pp. 1--14, 2024.

\bibitem{chen2023lifelong}
W.~Chen, Y.~Zhou, N.~Du, Y.~Huang, J.~Laudon, Z.~Chen, and C.~Cui, ``Lifelong language pretraining with distribution-specialized experts,'' in \emph{International Conferenceon Machine Learning}, 2023, pp. 5383--5395.

\bibitem{pan2015divide}
Y.~Pan, R.~Xia, J.~Yin, and N.~Liu, ``A divide-and-conquer method for scalable robust multitask learning,'' \emph{IEEE Transactions on Neural Networks and Learning Systems}, vol.~26, no.~12, pp. 3163--3175, 2015.

\bibitem{xia2021graph}
F.~Xia, K.~Sun, S.~Yu, A.~Aziz, L.~Wan, S.~Pan, and H.~Liu, ``Graph learning: A survey,'' \emph{IEEE Transactions on Artificial Intelligence}, vol.~2, no.~2, pp. 109--127, 2021.

\bibitem{lee2022augmentation}
N.~Lee, J.~Lee, and C.~Park, ``Augmentation-free self-supervised learning on graphs,'' in \emph{AAAI Conference on Artificial Intelligence}, vol.~36, no.~7, 2022, pp. 7372--7380.

\bibitem{liu2017accelerated}
N.~Liu, X.~Huang, and X.~Hu, ``Accelerated local anomaly detection via resolving attributed networks,'' in \emph{International Joint Conference on Artificial Intelligence}, 2017, pp. 2337--2343.

\bibitem{wu2024high}
H.~Wu, Y.~Wu, N.~Li, M.~Yang, J.~Zhang, M.~K. Ng, and J.~Long, \emph{Machine Learning}, vol. 113, no.~9, pp. 6247--6272, 2024.

\bibitem{mo2023multiplex}
Y.~Mo, Y.~Chen, Y.~Lei, L.~Peng, X.~Shi, C.~Yuan, and X.~Zhu, ``Multiplex graph representation learning via dual correlation reduction,'' \emph{IEEE Transactions on Knowledge and Data Engineering}, vol.~35, no.~12, pp. 12\,814--12\,827, 2023.

\bibitem{yu2019self}
W.~Yu, W.~Cheng, C.~Aggarwal, B.~Zong, H.~Chen, and W.~Wang, ``Self-attentive attributed network embedding through adversarial learning,'' in \emph{IEEE International Conference on Data Mining}, 2019, pp. 758--767.

\bibitem{gao2023addressing}
Y.~Gao, X.~Wang, X.~He, Z.~Liu, H.~Feng, and Y.~Zhang, ``Addressing heterophily in graph anomaly detection: A perspective of graph spectrum,'' in \emph{Proceedings of the ACM Web Conference}, 2023, pp. 1528--1538.

\bibitem{xiao2024Motif}
C.~Xiao, S.~Pang, W.~Tai, Y.~Huang, G.~Trajcevski, and F.~Zhou, ``Motif-consistent counterfactuals with adversarial refinement for graph-level anomaly detection,'' in \emph{Proceedings of the ACM SIGKDD Conference on Knowledge Discovery and Data Mining}, 2024, p. 3518–3526.

\bibitem{gao2023alleviating}
Y.~Gao, X.~Wang, X.~He, Z.~Liu, H.~Feng, and Y.~Zhang, ``Alleviating structural distribution shift in graph anomaly detection,'' in \emph{ACM International Conference on Web Search and Data Mining}, 2023, pp. 357--365.

\bibitem{shao2023learning}
M.~Shao, Y.~Lin, Q.~Peng, J.~Zhao, Z.~Pei, and Y.~Sun, ``Learning graph deep autoencoder for anomaly detection in multi-attributed networks,'' \emph{Knowledge-Based Systems}, vol. 260, p. 110084, 2023.

\bibitem{chen2023anomman}
L.-H. Chen, H.~Li, W.~Zhang, J.~Huang, X.~Ma, J.~Cui, N.~Li, and J.~Yoo, ``Anomman: Detect anomalies on multi-view attributed networks,'' \emph{Information Sciences}, vol. 628, pp. 1--21, 2023.

\bibitem{duan2023arise}
J.~Duan, B.~Xiao, S.~Wang, H.~Zhou, and X.~Liu, ``Arise: Graph anomaly detection on attributed networks via substructure awareness,'' \emph{IEEE Transactions on Neural Networks and Learning Systems}, vol.~35, no.~12, pp. 18\,172--18\,185, 2023.

\bibitem{hu2023samcl}
J.~Hu, B.~Xiao, H.~Jin, J.~Duan, S.~Wang, Z.~Lv, S.~Wang, X.~Liu, and E.~Zhu, ``Samcl: Subgraph-aligned multiview contrastive learning for graph anomaly detection,'' \emph{IEEE Transactions on Neural Networks and Learning Systems}, vol.~36, no.~1, pp. 1664--1676, 2023.

\bibitem{jin2021anemone}
M.~Jin, Y.~Liu, Y.~Zheng, L.~Chi, Y.-F. Li, and S.~Pan, ``Anemone: Graph anomaly detection with multi-scale contrastive learning,'' in \emph{ACM International Conference on Information \& Knowledge Management}, 2021, pp. 3122--3126.

\bibitem{ding2022data}
K.~Ding, Z.~Xu, H.~Tong, and H.~Liu, ``Data augmentation for deep graph learning: A survey,'' \emph{ACM SIGKDD Explorations Newsletter}, vol.~24, no.~2, pp. 61--77, 2022.

\bibitem{chen2020learning}
X.~Chen, S.~Chen, J.~Yao, H.~Zheng, Y.~Zhang, and I.~W. Tsang, ``Learning on attribute-missing graphs,'' \emph{IEEE Transactions on Pattern Analysis and Machine Intelligence}, vol.~44, no.~2, pp. 740--757, 2020.

\bibitem{jin2021heterogeneous}
D.~Jin, C.~Huo, C.~Liang, and L.~Yang, ``Heterogeneous graph neural network via attribute completion,'' in \emph{Proceedings of The Web Conference}, 2021, pp. 391--400.

\bibitem{tu2022initializing}
W.~Tu, S.~Zhou, X.~Liu, Y.~Liu, Z.~Cai, E.~Zhu, Z.~Changwang, and J.~Cheng, ``Initializing then refining: A simple graph attribute imputation network,'' in \emph{International Joint Conference on Artificial Intelligence}, 2022, pp. 3494--3500.

\bibitem{spinelli2020missing}
I.~Spinelli, S.~Scardapane, and A.~Uncini, ``Missing data imputation with adversarially-trained graph convolutional networks,'' \emph{Neural Networks}, vol. 129, pp. 249--260, 2020.

\bibitem{taguchi2021graph}
H.~Taguchi, X.~Liu, and T.~Murata, ``Graph convolutional networks for graphs containing missing features,'' \emph{Future Generation Computer Systems}, vol. 117, pp. 155--168, 2021.

\bibitem{gao2023handling}
Z.~Gao, Y.~Niu, J.~Cheng, J.~Tang, L.~Li, T.~Xu, P.~Zhao, F.~Tsung, and J.~Li, ``Handling missing data via max-entropy regularized graph autoencoder,'' in \emph{AAAI Conference on Artificial Intelligence}, 2023, pp. 7651--7659.

\bibitem{tan2023s2gae}
Q.~Tan, N.~Liu, X.~Huang, S.-H. Choi, L.~Li, R.~Chen, and X.~Hu, ``S2gae: Self-supervised graph autoencoders are generalizable learners with graph masking,'' in \emph{ACM International Conference on Web Search and Data Mining}, 2023, pp. 787--795.

\bibitem{wang2020inductive}
L.~Wang, B.~Zong, Q.~Ma, W.~Cheng, J.~Ni, W.~Yu, Y.~Liu, D.~Song, H.~Chen, and Y.~Fu, ``Inductive and unsupervised representation learning on graph structured objects,'' in \emph{International Conference on Learning Representations}, 2020.

\bibitem{yun2021neo}
S.~Yun, S.~Kim, J.~Lee, J.~Kang, and H.~J. Kim, ``Neo-gnns: Neighborhood overlap-aware graph neural networks for link prediction,'' in \emph{Advances in Neural Information Processing Systems}, 2021, pp. 13\,683--13\,694.

\bibitem{zhu2021neural}
Z.~Zhu, Z.~Zhang, L.-P. Xhonneux, and J.~Tang, ``Neural bellman-ford networks: A general graph neural network framework for link prediction,'' in \emph{Advances in Neural Information Processing Systems}, 2021, pp. 29\,476--29\,490.

\bibitem{guo2023linkless}
Z.~Guo, W.~Shiao, S.~Zhang, Y.~Liu, N.~V. Chawla, N.~Shah, and T.~Zhao, ``Linkless link prediction via relational distillation,'' in \emph{International Conferenceon Machine Learning}, 2023, pp. 12\,012--12\,033.

\bibitem{zhao2022learning}
T.~Zhao, G.~Liu, D.~Wang, W.~Yu, and M.~Jiang, ``Learning from counterfactual links for link prediction,'' in \emph{International Conferenceon Machine Learning}, 2022, pp. 26\,911--26\,926.

\bibitem{liu2024self}
M.~Liu, K.~Liang, Y.~Zhao, W.~Tu, S.~Zhou, X.~Gan, X.~Liu, and K.~He, ``Self-supervised temporal graph learning with temporal and structural intensity alignment,'' \emph{IEEE Transactions on Neural Networks and Learning Systems}, vol.~36, no.~4, pp. 6355--6367, 2024.

\bibitem{wang2019learning}
L.~Wang, W.~Yu, W.~Wang, W.~Cheng, W.~Zhang, H.~Zha, X.~He, and H.~Chen, ``Learning robust representations with graph denoising policy network,'' in \emph{IEEE International Conference on Data Mining}, 2019, pp. 1378--1383.

\end{thebibliography}
